\pdfoutput=1

\documentclass[11pt]{article}

\usepackage[final]{acl}

\usepackage{times}
\usepackage{latexsym}

\usepackage[T1]{fontenc}

\usepackage[utf8]{inputenc}

\usepackage{microtype}

\usepackage{inconsolata}

\usepackage{graphicx}
\usepackage{multirow}
\usepackage{makecell}
\usepackage{color}
\usepackage{colortbl}
\usepackage{pifont}
\usepackage{subfig}
\usepackage{array}
\usepackage{amssymb}
\usepackage{amsmath}
\usepackage[ruled, vlined]{algorithm2e}
\newcommand{\tabincell}[2]{\begin{tabular}{@{}#1@{}}#2\end{tabular}}

\title{Knowledge-Guided Dynamic Modality Attention Fusion Framework for Multimodal Sentiment Analysis}

\newcommand*\samethanks[1][\value{footnote}]{\footnotemark[#1]}

\newcommand \footnoteONLYtext[1]
{
\let \mybackup \thefootnote
\let \thefootnote \relax
\footnotetext{#1}
\let \thefootnote \mybackup
\let \mybackup \imareallyundefinedcommand
}

\author{Xinyu Feng \quad Yuming Lin\samethanks[1] \quad Lihua He \quad You Li \quad Liang Chang \quad Ya Zhou \\
  Guangxi Key Laboratory of Trusted Software, \\ Guilin University of Electronic Technology, Guangxi, China \\
}

\begin{document}
\maketitle
\makeatletter\def\Hy@Warning#1{}\makeatother
\footnoteONLYtext{*Corresponding author. Email: ymlin@guet.edu.cn.}
\begin{abstract}
  Multimodal Sentiment Analysis (MSA) utilizes multimodal data to infer the users' sentiment. Previous methods focus on equally treating the contribution of each modality or statically using text as the dominant modality to conduct interaction, which neglects the situation where each modality may become dominant. In this paper, we propose a Knowledge-Guided Dynamic Modality Attention Fusion Framework (KuDA) for multimodal sentiment analysis. KuDA uses sentiment knowledge to guide the model dynamically selecting the dominant modality and adjusting the contributions of each modality. In addition, with the obtained multimodal representation, the model can further highlight the contribution of dominant modality through the correlation evaluation loss. Extensive experiments on four MSA benchmark datasets indicate that KuDA achieves state-of-the-art performance and is able to adapt to different scenarios of dominant modality.\footnote{Our code is publicly available at \url{https://github.com/MKMaS-GUET/KuDA}}
\end{abstract}

\section{Introduction}
\label{sec:Introduction}
\begin{figure}[ht]
  \includegraphics[width=\linewidth]{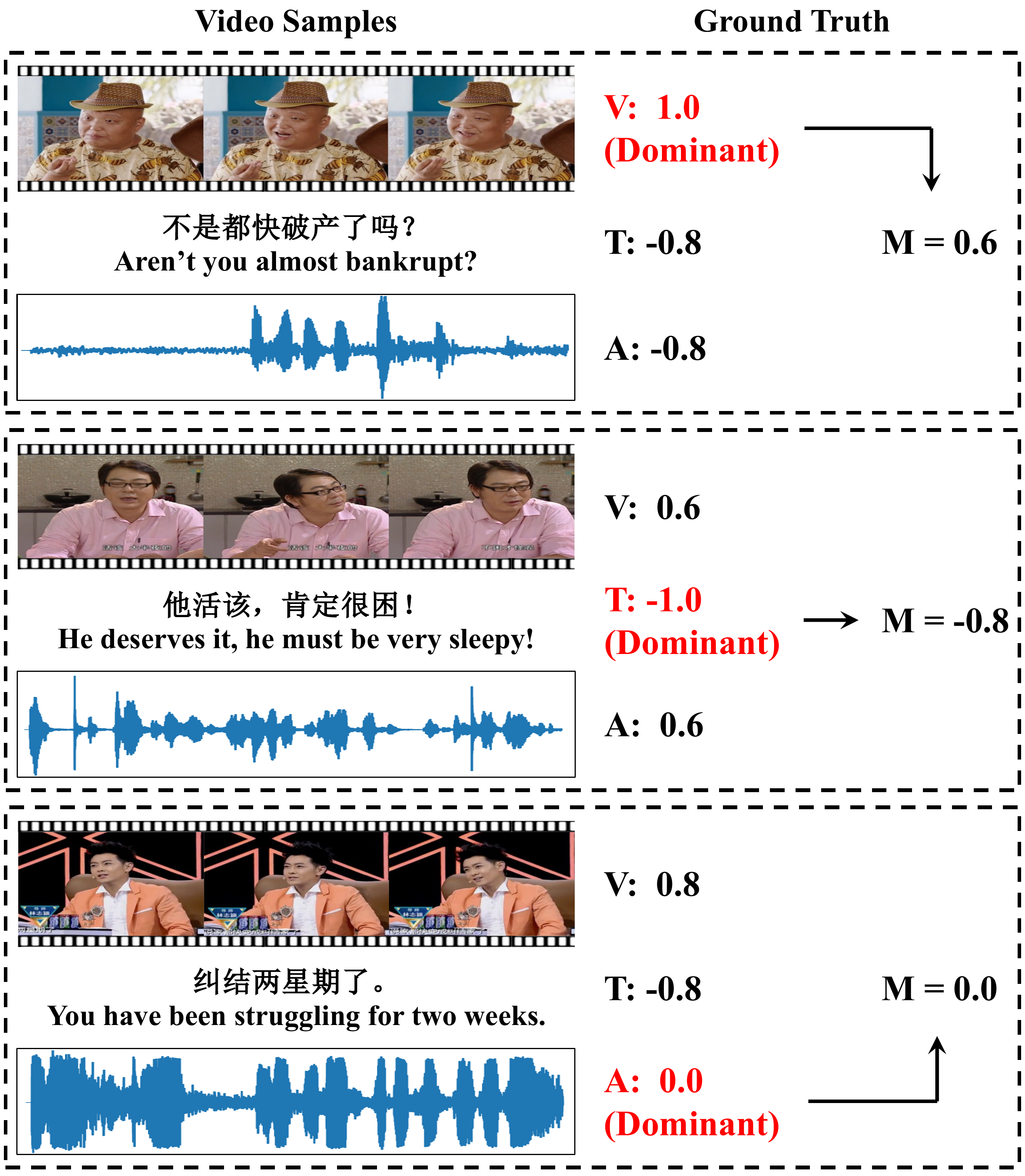}
  \caption{Three video samples with vision, text, or audio as the dominant modality from CH-SIMS dataset. In the Ground Truth, V, T and A denote the vision, text and audio, respectively. M is the overall sentiment label of the video sample. The values of all labels are from -1 (negative) to 1 (positive).}
  \label{fig:motivation}
\end{figure}

Since users' sentiment expressions are reflected in multiple modalities on social media, multimodal sentiment analysis (MSA) has garnered rising attention in recent years. It aims to mine and comprehend the sentiments of online videos \citep{soleymani2017survey,zeng2021contribution,kaur2022multimodal}. Most recent MSA methods can be grouped into two categories: ternary symmetric-based methods \citep{zadeh2017tensor,liu2018efficient,zadeh2018memory,tsai2019multimodal,hazarika2020misa,yu2021learning,sun2022cubemlp,huang2024tmbl} and text center-based methods \citep{han2021bi,han2021improving,wang2022cross,li2022amoa,lin2022multimodal,wang2023tetfn,ZhangWYL0Y23}. Ternary symmetric-based methods focus on equally treating the contribution of each modality and modeling the bidirectional relationship of all modality pairs. Text center-based methods focus on using text as the dominant modality to guide the vision and audio modalities to interact with it to adjust the contributions of different modalities properly. Thus, both ternary symmetric-based methods and text center-based methods consider the distribution of importance among modalities to be static and fix the dominant modality.

However, as shown in Figure~\ref{fig:motivation}, we discovered that in certain situations, vision, text, or audio could be the dominant modality respectively. With the first sample in Figure~\ref{fig:motivation}, since the vision label is more consistent with the overall sentiment label, it is dominant. Based on an investigation of commonly used datasets in MSA, we find that these situations are not uncommon. Detailed statistics and analysis can be found in Appendix~\ref{sec:DataInvestigationandStatistics}. Thus, when the dominant modality is not fixed, the ternary symmetric-based methods cannot effectively adapt to the situation where any modality is dominant because they do not consider the differences of importance between modalities. The text center-based methods statically set text as the dominant modality, and when other modalities are dominant, the model's attention is distracted by the text.

In this paper, we propose a \textbf{K}nowledge-G\textbf{u}ided \textbf{D}ynamic Modality \textbf{A}ttention Fusion Framework (KuDA), which improves the model performance and makes it adaptable to more complex and wider scenarios by dynamically selecting the dominant modality and adjusting the contributions of each modality according to different samples. Specifically, KuDA first uses the BERT model and two Transformer Encoders to extract semantic features of text, vision, and audio modalities. Then, KuDA performs sentiment knowledge injection and sentiment ratio conversion by the adapters and decoders, which can extract sentiment clues and guide KuDA in selecting the dominant modality further. Next, the dynamic attention fusion module is designed to capture similar sentiment information and gradually adjusts the attention weights between modalities by interacting sentiment knowledge with different levels of multimodal features. Based on the correlation evaluation between the multimodal features and the unimodal features, we use Noise-Contrastive Estimation \citep{gutmann2010noise} to highlight the contribution of the dominant modality further. Finally, KuDA predicts the sentiment score through a multilayer perceptron.

The main contributions of our work can be summarized as follows:
\begin{itemize}
  \item[$\bullet$] We propose KuDA, a Knowledge-Guided Dynamic Modality Attention Fusion Framework for multimodal sentiment analysis, which improves the performance by dynamically selecting the dominant modality, making model adaptable to complex and wide scenarios.
  \item[$\bullet$] We design a Dynamic Attention Fusion module, which utilizes sentiment knowledge to guide different levels of multimodal features and achieves dynamic fusion by adjusting the contribution of each modality.
  \item[$\bullet$] Extensive experiments on four MSA datasets show that KuDA achieves state-of-the-art performance. We further analyze the experimental results to prove the effectiveness of KuDA.
\end{itemize}

\section{Related Work}
\label{sec:RelatedWork}
In this section, we briefly overview related work in ternary symmetric-based methods and text center-based methods.

\begin{figure*}[ht]
  \includegraphics[width=\linewidth]{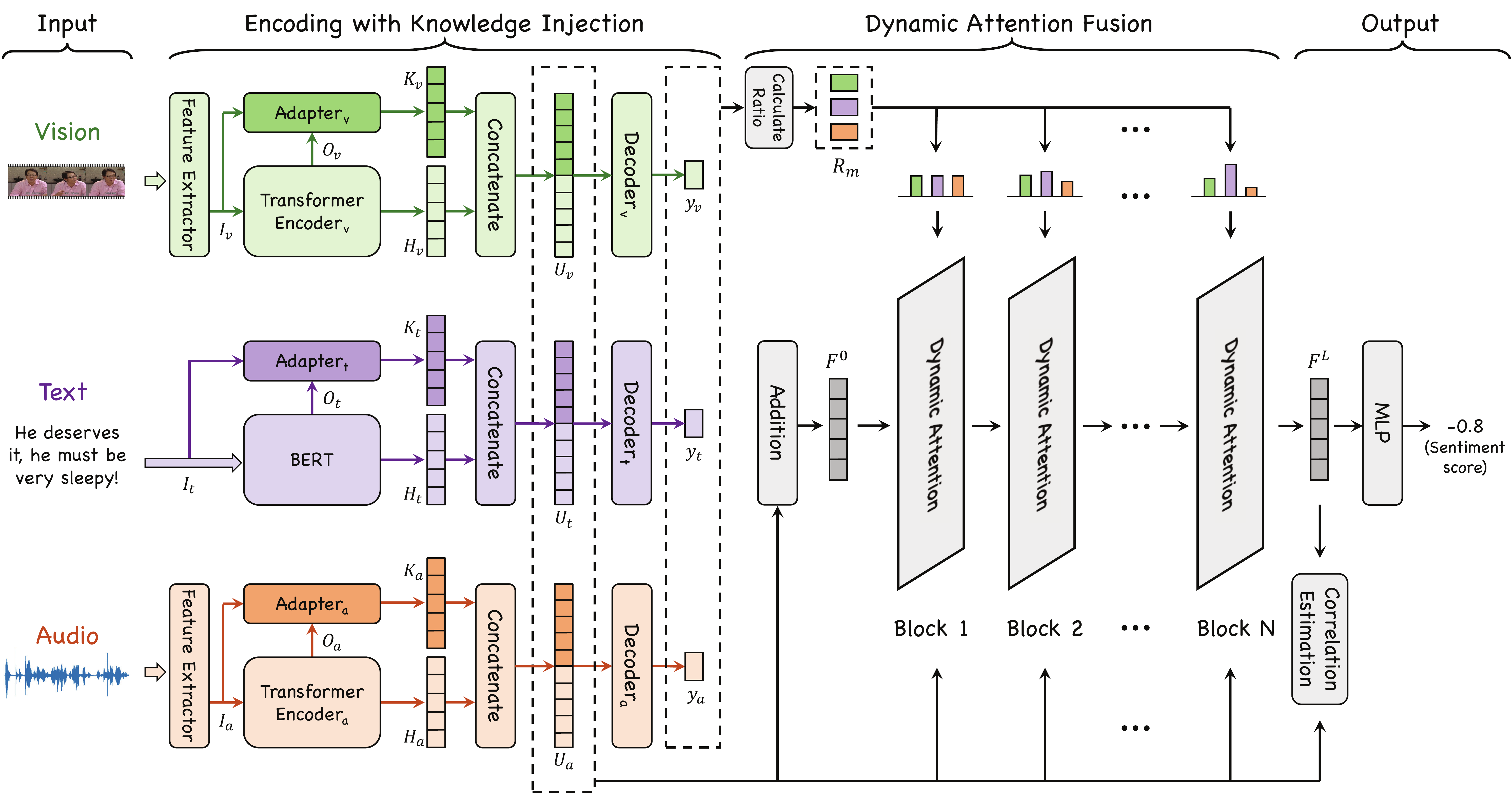}
  \caption{The overall architecture of KuDA. The green, purple, orange, and gray modules represent the relevant operations of vision, text, audio and multimodal fusion.}
  \label{fig:overallmodel}
\end{figure*}

\textbf{Ternary symmetric-based methods.} Previous works in this category have primarily focused on modeling bidirectional relationships in each modality pair and treating the contribution of each modality equally. For instance, some researchers designed the fusion architectures of tensor-based \citep{zadeh2017tensor,liu2018efficient}, LSTMs-based \citep{zadeh2018memory} and mlp-based \citep{sun2022cubemlp} to capture the commonalities between modalities. While \citet{tsai2019multimodal} designed MuLT, which is a transformer-based model, to find similar information between each modality pair. Some studies \citep{hubert2018learning,hazarika2020misa,YangDisentangled} project each modality into two subspaces, separately learning the commonalities and characteristics of the modalities to aid the fusion process. Multi-label strategy \citep{yu2021learning} and contrastive learning \citep{mai2022hybrid} are introduced to improve the quality of unimodal features. In addition, \citet{huang2024tmbl} designed TMBL, which can handle bimodal and trimodal features to capture and utilize bound modal features.

\textbf{Text center-based methods.} Previous works in this category mainly focus on improving the quality of fuse representation through the text modality to guide the vision and audio modalities. For example, \citet{delbrouck2020transformer}, \citet{han2021bi} and \citet{wang2023tetfn} use the transformer-based method to integrate similar information from other modalities by text modality. While some researchers \citep{rahman2020integrating,wang2022cross} enhance the text representations by integrating vision and audio information into a pretrained language model. Moreover, to decrease the potential sentiment-irrelevant and conflicting information, some studies reduce additional noise by maximizing mutual information \citep{han2021improving} or adaptive representation learning \citep{ZhangWYL0Y23}. Meanwhile, contrastive learning is introduced to learn invariant and similar information between text with other modalities \citep{li2022amoa,lin2022multimodal}.

Despite the promising results achieved for MSA, the above approaches treat each modality in a balanced way or statically set text as the dominant modality. This causes the approach to be distracted by the secondary modalities, hindering it from dynamically adjusting to diverse scenarios where the dominant modality varies, which limits the performance of MSA.

\section{Methodology}
\label{sec:Method}

\subsection{Overall Architecture}
\label{sec:OverallArchitecture}
As shown in Figure~\ref{fig:overallmodel}, which shows the overall workflow of KuDA. Specifically, KuDA first extracts unimodal low-level features from the raw multimodal input. Then, the adapters and encoders extract unimodal high-level features and learn sentiment knowledge simultaneously. We utilize the decoders to predict the unimodal sentiments and convert them into sentiment ratios to guide dynamic fusion. Next, we designed a dynamic attention fusion module, which selects the dominant modality according to different scenarios and dynamically adjusts attention weights with sentiment ratios and knowledge representations. Finally, the multimodal representation is used to conduct the MSA task by multilayer perceptron (MLP) and estimate correlation with knowledge representations.

In addition, to guide the model in adjusting the attention weights by sentiment knowledge, KuDA adopts a two-stage training method.

\subsection{Problem Definition and Notations}
\label{sec:ProblemDefinitionandNotations}
In MSA task, the input data consists of text $\left(t\right)$, vision $\left(v\right)$ and audio $\left(a\right)$ modalities. The sequences of three modalities are represented as triplet $\left(I_{t}, I_{v}, I_{a}\right)$, which include $I_{t} \in \mathbb{R}^{T_{t} \times d_{t}}$, $I_{v} \in \mathbb{R}^{T_{v} \times d_{v}}$, and $I_{a} \in \mathbb{R}^{T_{a} \times d_{a}}$, where $T_{m}, m \in \left\{t, v, a\right\}$ is the sequence length and $d_{m}$ represents the vector dimension. The prediction is the sentiment score $\hat{y}$, which is a discrete value between [-1, 1] and [-3, 3], with values greater than, equal to, and less than 0 representing positive, neutral, and negative, respectively.

\subsection{Encoding with Knowledge Injection}
\label{sec:KnowledgeInjectionandEncoder}
We encode the input of each modality $I_{m\in \left \{ t,v,a \right \} }$ into the global semantic representations $H_{m}\in \mathbb{R}^{T_{m} \times d_{m}}$ and the knowledge-sentiment representations $K_{m}\in \mathbb{R}^{T_{m} \times d_{m}}$ via the pretrained encoders and adapters, respectively.

\textbf{Global semantic representations.} For the text modality, to effectively extract from low-level to high-level text semantic information and facilitate subsequent knowledge inject with Adapter, we use the BERT \citep{kenton2019bert} to encode the input sentences $I_{t}$, and extract the hidden state of the last layer as the global semantic representation $H_{t}$:

\begin{figure}[htp]
  \includegraphics[width=\columnwidth]{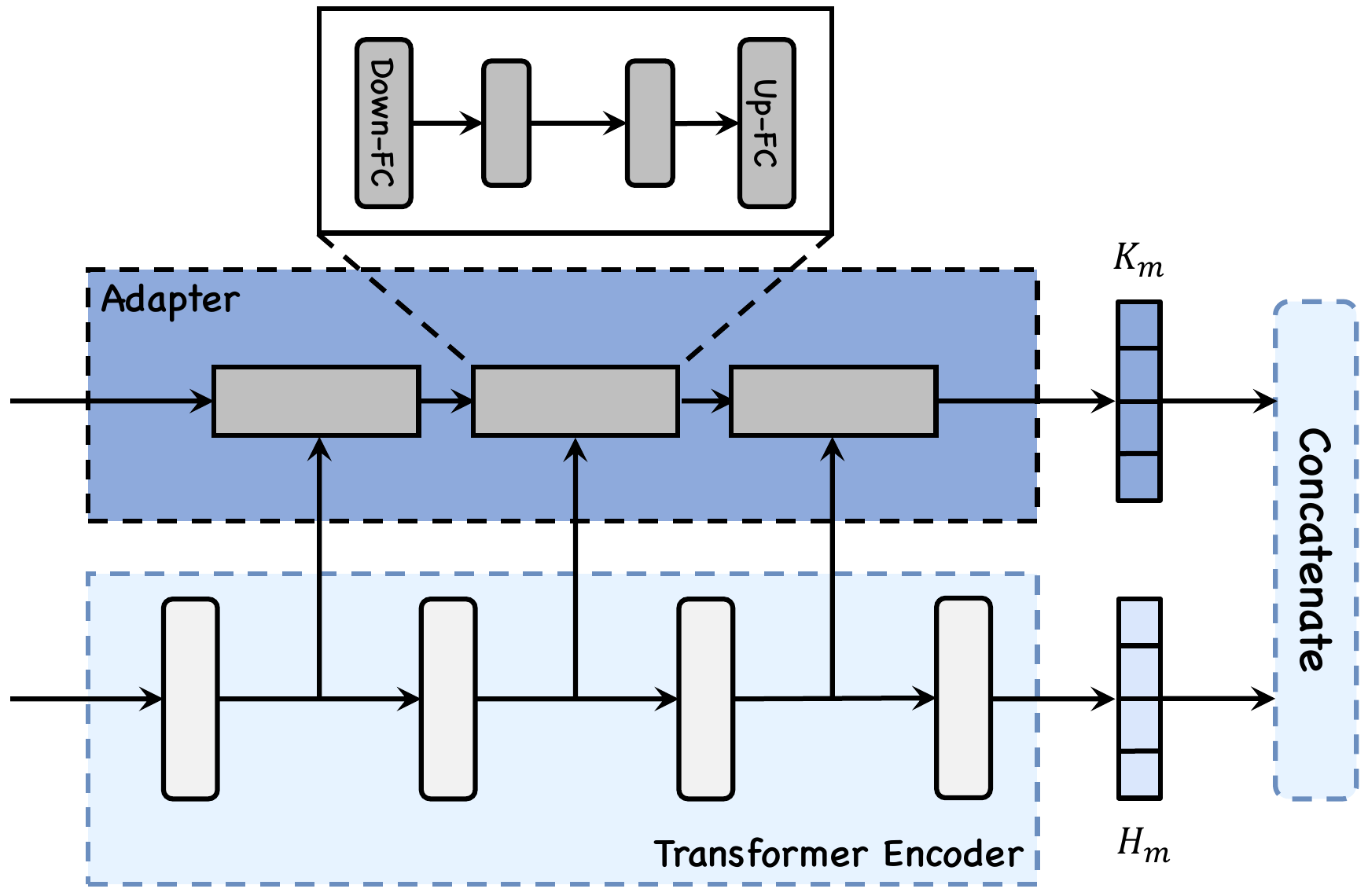}
  \caption{The architecture of the BERT, transformer encoder, adapter, and their connection. The vertical gray rectangles denote the Transformer Encoder Layer. Down-FC and Up-FC represent the fully connected layers (FC) used to decrease and increase the dimension.}
  \label{fig:adapter}
\end{figure}

\begin{equation}
  \label{eq:bertencoder}
  H_{t}, O_{t} = \text{BERT} \left ( I_{t} ;  \theta^{\text{BERT}} \right )
\end{equation}

\noindent where $O_{t}$ denotes the hidden states except the last layers. Similar to text modality, for the vision and audio modalities, we use stacked transformer encoder layers \citep{vaswani2017attention} to capture global semantic representations $H_{m}, m \in \left \{ v, a \right \}$:

\begin{equation}
  \label{eq:tfencoder}
  H_{m}, O_{m} = \text{Encoder}_{m} \left ( I_{m} ;  \theta^{\text{encoder}}_{m} \right )
\end{equation}

\noindent Since the obtained $O_{m}, m \in \left \{ t, v, a \right \}$ mainly include the general knowledge, they are input into the Adapter to inject the sentiment information.

\textbf{Knowledge-sentiment representations.} Since the adapter is commonly used to enhance the knowledge cognition of pretrained language models \citep{wei2021knowledge}, we use it to inject unimodal sentiment knowledge. In KuDA, the adapter is plugged outside of the encoder and stacked with identical blocks, and its detailed architecture is shown in Figure~\ref{fig:adapter}. We connect each transformer encoder layer of the vision and audio modalities to the Adapter. To prevent information redundancy, we select part of the intermediate layers of BERT to connect. For the first adapter block, we take the unimodal features $I_{m\in \left \{ t,v,a \right \} }$ as one of the inputs. The output of adapter is denoted as knowledge-sentiment representation $K_{m}, m \in \left \{ t,v,a \right \}$:

\begin{equation}
  \label{eq:adapter}
  K_{m} = \text{Adapter}_{m} \left ( I_{m}, O_{m} ;  \theta^{\text{adapter}}_{m} \right ) \in \mathbb{R}^{T_{m} \times d_{m}}
\end{equation}

\noindent where $\theta^{\text{Adapter}}_{m}$ is denoted the pretrained parameters of the adapter of $m$ modality.

\textbf{Unimodal sentiment scores.} We combine the above two representations to obtain the knowledge-enhanced representation of each modality $U_{m}, m \in \left \{t, v, a \right \}$. Then, we use this representation to predict the unimodal sentiment score $\hat{y}_m$ by a decoder consisting of MLP:

\begin{align}
  U_{m} &= \left [ K_{m} ; H_{m} \right ]\in \mathbb{R}^{T_{m} \times 2d_{m}} \label{eq:knowenhance} \\
  \hat{y}_m &= \text{Decoder}_{m}\left ( U_{m} ; \theta^{\text{decoder}}_{m}\right ) \label{eq:decoder}
\end{align}

\noindent where $\left [ \cdot ; \cdot \right ]$ denotes the concatenation. Due to the difference between the unimodal and multimodal sentiment scores can indicate the effective amount of information provided by the corresponding modality, so we use the sentiment scores $\hat{y}_m$ to guide attention weights further.

\subsection{Dynamic Attention Fusion}
\label{sec:DynamicAttentionFusion}

\subsubsection{Unimodal Sentiment Ratio}
\label{InteractionofUnimodalSentimentRatio}
Since the difference in sentiment score of unimodal $y_{m}$ and multimodal $y$ is inversely proportional to the weight of unimodal, we choose the inverse proportional function $\exp\left ( -kx \right )$ and normalization operation, and utilize the ground truth of MSA $y$ to convert the unimodal sentiment score $\hat{y}_{m}$ into sentiment ratio $R_{m}, m \in \left \{t, v, a \right \}$ during training to guide subsequent dynamic fusion:

\begin{equation}
  \label{eq:unisentiratio}
  \begin{aligned}
    D_{m} &= \exp\left ( -k\left | \hat{y}_{m} - y \right |^{2} \right ) \\
    R_{m} &= \frac{D_{m}}{D_{t}+D_{v}+D_{a}}
  \end{aligned}
\end{equation}

\noindent where $k$ denotes the function's slope, which can scale the sentiment ratio. Due to the model effectively learned how to adjust the contribution between modalities during training, we fixed the sentiment ratio to 1 during the test stage to highlight the model's ability to adjust weights.

\subsubsection{Dynamic Attention Block}
\label{DynamicFusionBlock}
In order to unify the length and dimension axis of the unimodal knowledge-enhanced representation $U_{m} \in \mathbb{R}^{T_{m} \times 2d_{m}}$ for multimodal fusion, we utilize three projectors to obtain the updated knowledge-enhanced representation of each modality $\overline{U}_{m}, m \in \left \{t, v, a \right \}$. In addition, due to any one or more of the text, vision, and audio modalities may become the dominant modality, we first sum the obtained representations $\overline{U}_{m}$ as the input of the first dynamic attention block $F^{0}$:

\begin{figure}[htp]
  \includegraphics[width=\columnwidth]{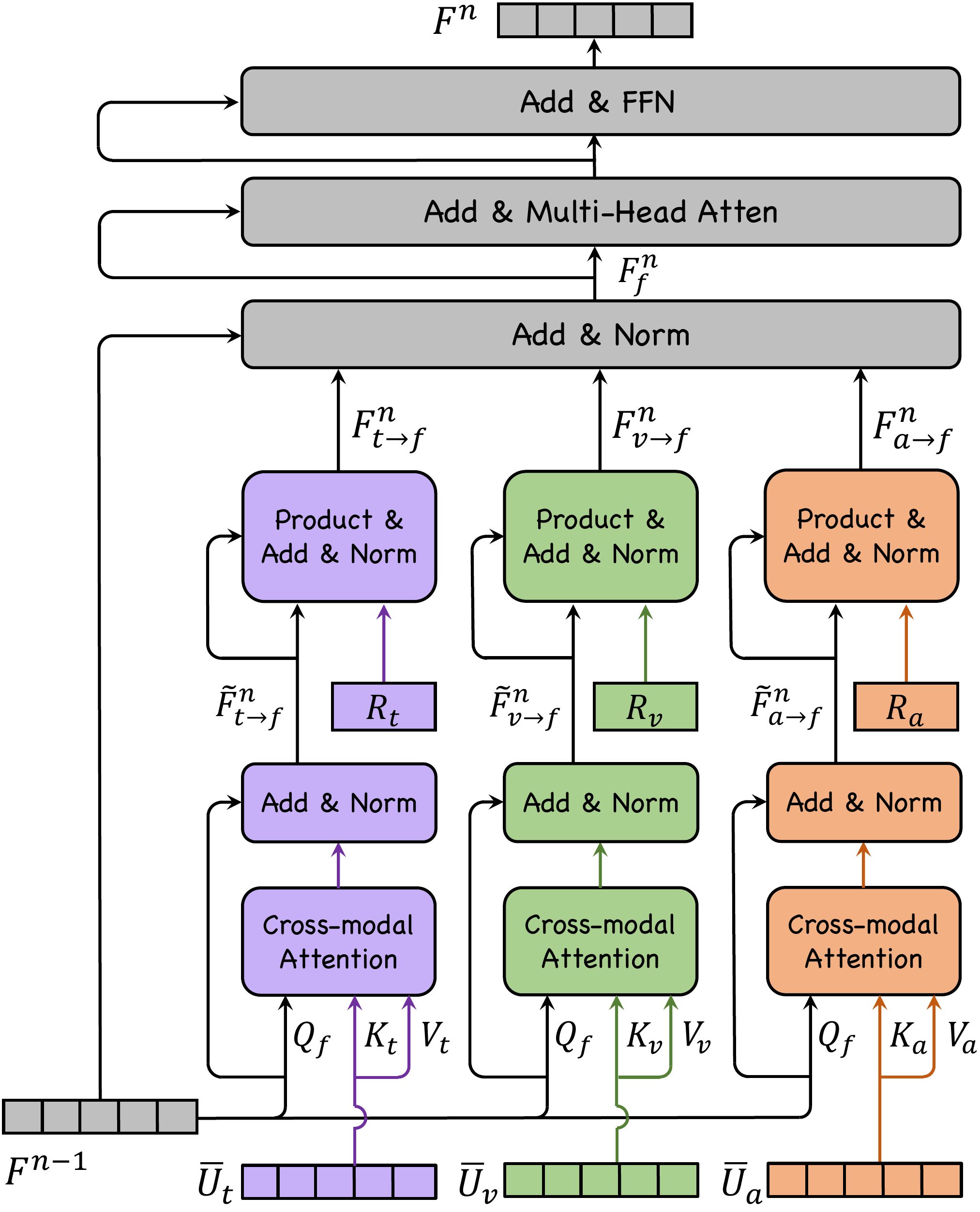}
  \caption{Architecture of a single dynamic attention block. The purple, green, and orange represent the interaction of multimodal representation with text, vision, and audio modalities, respectively.}
  \label{fig:dynamicfusion}
\end{figure}

\begin{equation}
  \label{eq:firstblock}
  \begin{aligned}
    \overline{U}_{m} &= \text{Projector}_{m}\left ( U_{m} \right ) \in \mathbb{R}^{T_{f} \times d_{f}} \\
    F^{0} &= \overline{U}_{t} + \overline{U}_{v} + \overline{U}_{a} \in \mathbb{R}^{T_{f} \times d_{f}}
  \end{aligned}
\end{equation}

\noindent where $\text{Projector}_{m}$ consists of two linear layers, $T_{f}$ and $d_{f}$ denote length and dimension in fusion stage.

We then stack the dynamic attention blocks to form a pipeline, which is shown in Figure~\ref{fig:overallmodel}. At the same time, we use the output of the prior block $F^{n-1}$, the knowledge-enhanced representation $\overline{U}_{m}$, and the sentiment ratio $R_{m}, m \in \left \{t, v, a \right \}$ as the input of the next block and obtain its output $F^{n}$:

\begin{equation}
  \label{eq:dynamicattentionblock}
  F^{n} = \text{DAB} \left ( F^{n-1}, \overline{U}_{m}, R_{m} \right ) \in \mathbb{R}^{T_{f} \times d_{f}}
\end{equation}

\noindent where DAB is the dynamic attention block. This is because the representations $\overline{U}_{m}$ have richer sentiment clues, and the sentiment ratios $R_{m}$ can indicate the contribution of each modality. Finally, we take the output of the last block as the final multimodal representation $F^{L} \in \mathbb{R}^{T_{f} \times d_{f}}$ and use it to conduct the MSA task.

As shown in Figure~\ref{fig:dynamicfusion}, to adjust the weights of each modality, we designed a dynamic attention block. Specifically, we first introduce the cross-modal attention (CAttn), which gradually determines the dominant modality by capturing the amount of similar information between unimodal representations $\overline{U}_{m}, m \in \left \{ t,v,a \right \}$ and multimodal representation $F^{n-1}$. Since $Q$ of attention is used to specify the location of attention, we take the multimodal features as $Q$, and the unimodal features as $K$ and $V$, and perform the Layer Norm (LN):

\begin{equation}
  \label{eq:unimulti}
  \tilde{F}_{m\to f}^{n} = \text{LN}\left ( F^{n-1} + \text{CAttn}\left ( F^{n-1}, \overline{U}_{m}, \overline{U}_{m} \right ) \right )
\end{equation}

Next, since the sentiment ratio $R_{m}$ can further guide the dynamic fusion, we multiply it with the middle representation $\tilde{F}_{m\to f}^{n}, m \in \left \{t, v, a \right \}$. We then sum the obtained representations and the multimodal representation of input $F^{n-1}$ to fine-tune the contributions of different modalities:

\begin{align}
  \label{eq:fusion1} F_{m\to f}^{n} &= \text{LN}\left ( \tilde{F}_{m\to f}^{n} + \left ( R_{m} \times \tilde{F}_{m\to f}^{n} \right ) \right ) \\
  \label{eq:fusion2} F_{f}^{n} &= F^{n-1} + \text{LN}\left ( \sum_{m\in \left \{ t,v,a \right \} } F_{m\to f}^{n} \right )
\end{align}

At last, we input the $F_{f}^{n}$ into the multi-head attention and feedforward neural network to get the output of dynamic attention block $F^{n} \in \mathbb{R}^{T_{f} \times d_{f}}$.

\subsection{Output and Training Objectives}
\label{sec:TrainingProcedure}
To further improve the utilization of the dominant modality, we estimate the correlation of the multimodal representation $F^{L}$ and the unimodal representations $\overline{U}_{m}, m \in \left \{ t,v,a \right \}$ through the Contrastive Predictive Coding \citep{oord2018representation}, and integrated it into the Noise-Contrastive Estimation framework \citep{gutmann2010noise} to form the correlation estimation (CE) loss $\mathcal{L}_{cor}$:

\begin{equation}
  \label{eq:cor_loss}
    \mathcal{L}_{cor} = \sum_{m\in \left \{ t, v, a \right \} } \mathcal{L}_{\text{NCE}}\left ( F^{L}, \overline{U}_{m} \right )
\end{equation}

In the Output, we input the representation $F^{L}$ into a MLP to predict sentiment score $\hat{y}$. Given the predictions $\hat{y}$ and the ground truth $y$, we calculate the MSA task loss $\mathcal{L}_{reg}$ by mean absolute error. Finally, we training KuDA by the union loss $\mathcal{L}_{task}$:

\begin{align}
  \label{eq:prediction} \hat{y} &= \text{MLP}\left( \text{Mean}\left ( F^{L} \right ) \right) \\
  \label{eq:loss_reg} \mathcal{L}_{reg} &= \frac{1}{N}\sum_{i=1}^{N}\left|y_i-{\hat{y}}_i\right| \\
  \label{eq:loss_task} \mathcal{L}_{task} &= \mathcal{L}_{reg}+\alpha\mathcal{L}_{cor}
\end{align}

\begin{table}[ht]\small
  \renewcommand{\arraystretch}{1.1}
  \setlength{\tabcolsep}{4.1pt}
  \centering
  \begin{tabular}{lccccc}
    \hline
    Dataset    & \#Train & \#Valid & \#Test & \#Total & Language \\
    \hline
    CH-SIMS    & 1368   & 456    & 457   & 2281   & Chinese   \\
    CH-SIMSv2  & 2722   & 647    & 1034  & 4403   & Chinese   \\
    MOSI       & 1284   & 229    & 686   & 2199   & English   \\
    MOSEI      & 16326  & 1871   & 4659  & 22856  & English   \\
    \hline
  \end{tabular}
  \caption{The statistics of CH-SIMS, CH-SIMSv2, MOSI and MOSEI.}
  \label{tab:Datasets_split}
\end{table}

\noindent where $\text{Mean}\left ( \cdot  \right )$ denotes the average operation in length axis. $\alpha$ is a parameter that balances the contribution of different losses. The specific training algorithm of KuDA can be found in Appendix~\ref{sec:TrainingProcess}.

\section{Experiments}
\label{sec:Experiments}

\subsection{Datasets and Evaluation Metrics}
\label{sec:DatasetsandEvaluationMetrics}

We conduct experiments on four publicly benchmark datasets of MSA, including CH-SIMS \citep{yuetal2020ch}, CH-SIMSv2 \citep{liu2022make}, MOSI \citep{zadeh2016multimodal} and MOSEI \citep{zadeh2018multimodal}. The statistic details of four datasets are shown in Table~\ref{tab:Datasets_split}.

Following previous works \citep{hazarika2020misa,yu2021learning,ZhangWYL0Y23}, we used the accuracy of 3-class (Acc-3) and 5-class (Acc-5) on CH-SIMS and CH-SIMSv2, the accuracy of 7-class (Acc-7) on MOSI and MOSEI, and the accuracy of 2-class (Acc-2), Mean Absolute Error (MAE), Pearson Correlation (Corr), and F1-score (F1) on all datasets. Moreover, on MOSI and MOSEI, Acc-2 and F1 used two calculation ways: negative/non-negative (has-0) and negative/positive (non-0). Except for MAE, higher values indicate better performance for all metrics.

\subsection{Baselines}
\label{sec:Baselines}
To validate the KuDA's performance, we conduct a fair comparison with several competitive and state-of-the-art (SOTA) baselines, including the ternary symmetric-based methods: TFN \citep{zadeh2017tensor}, LMF \citep{liu2018efficient}, MuLT \citep{tsai2019multimodal}, MISA \citep{hazarika2020misa}, Self-MM \citep{yu2021learning}, CubeMLP \citep{sun2022cubemlp} and TMBL \citep{huang2024tmbl}, and the text center-based methods: MMIM \citep{han2021improving}, BBFN \citep{han2021bi}, CENet \citep{wang2022cross}, TETFN \citep{wang2023tetfn} and ALMT \citep{ZhangWYL0Y23}.

\begin{table}[ht]\scriptsize
  \renewcommand{\arraystretch}{1.1}
  \setlength{\tabcolsep}{2.5pt}
  \centering
  \begin{tabular}{l|cccc}
    \hline
    Descriptions & CH-SIMS & CH-SIMSv2 & MOSI & MOSEI \\
    \hline
    Batch Size                      & 32       & 32       & 32    & 64    \\
    Initial Learning Rate           & 3e-5     & 3e-5     & 3e-5  & 4e-5  \\
    Knowledge Injection of Bert     & 3,6,9,11 & 3,6,9,11 & 6,9   & 6,9   \\
    Dynamic Transformer Block       & 3        & 3        & 2     & 4     \\
    Vector Dimension $F^{i}$        & 256      & 256      & 256   & 256   \\
    k                               & 0.1      & 0.3      & 2.0   & 5.0   \\
    $\alpha$                        & 0.01     & 0.01     & 0.01  & 0.1   \\
    Epochs                          & 50       & 50       & 50    & 50    \\
    Optimizer                       & AdamW    & AdamW    & AdamW & AdamW \\
    \hline
  \end{tabular}
  \caption{Hyper-parameters settings on different datasets.}
  \label{tab:Hyperparameters}
\end{table}

\subsection{Experimental Settings}
\label{sec:ExperimentalSettings}
To ensure fairness with other baselines, we follow recent competitive and SOTA methods to set the proposed method. The training stages consist of the unimodal stage and the multimodal stage. In addition, we adopt BERT to extract the features of text modality where ``bert-base-chinese''\footnote{https://huggingface.co/bert-base-chinese} is employed for CH-SIMS and CH-SIMSv2 and ``bert-base-uncased''\footnote{https://huggingface.co/bert-base-uncased} is employed for MOSI and MOSEI. In the vision and audio modalities, we directly use the features provided by the original datasets. Moreover, we develop the KuDA using a single NVIDIA RTX 3090 GPU for all datasets. The detailed setting of the best hyper-parameters can be referred to in Table \ref{tab:Hyperparameters}.

\begin{table*}[ht]\small
  \setlength{\tabcolsep}{6pt}
  \centering
  \begin{tabular}{l|cccccc|cccccc}
    \hline
    \multicolumn{1}{l|}{\multirow{2}{*}{\textbf{Methods}}} & \multicolumn{6}{c|}{\textbf{CH-SIMS}} & \multicolumn{6}{c}{\textbf{CH-SIMSv2}} \\
    {} & \textbf{MAE} & \textbf{Corr} & \textbf{Acc-5} & \textbf{Acc-3} & \textbf{Acc-2} & \textbf{F1} & \textbf{MAE} & \textbf{Corr} & \textbf{Acc-5} & \textbf{Acc-3} & \textbf{Acc-2} & \textbf{F1} \\
    \hline
    TFN${}^\dagger$     & 0.432 & 0.591 & 39.30 & 65.12 & 78.38 & 78.62 & 0.303 & 0.707 & 52.55 & 72.21 & 80.14 & 80.14 \\
    LMF${}^\dagger$     & 0.441 & 0.576 & 40.53 & 64.68 & 77.77 & 77.88 & 0.367 & 0.557 & 47.79 & 64.90 & 74.18 & 73.88 \\
    MulT${}^\dagger$    & 0.453 & 0.564 & 37.94 & 64.77 & 78.56 & 79.66 & 0.291 & 0.738 & 54.81 & 73.19 & 80.68 & 80.73 \\
    BBFN${}^*$          & 0.430 & 0.564 & 40.92 & 61.05 & 78.12 & 77.88 & 0.300 & 0.708 & 53.29 & 71.47 & 78.53 & 78.41 \\
    Self-MM${}^\dagger$ & 0.425 & 0.595 & 41.53 & 65.47 & 80.04 & 80.44 & 0.311 & 0.695 & 52.77 & 72.61 & 79.69 & 79.76 \\
    CubeMLP${}^*$       & 0.419 & 0.593 & 41.79 & 65.86 & 77.68 & 77.59 & 0.334 & 0.648 & 52.90 & 71.95 & 78.53 & 78.53 \\
    CENet${}^\dagger$   & 0.471 & 0.534 & 33.92 & 62.58 & 77.90 & 77.53 & 0.310 & 0.699 & 53.04 & 73.10 & 79.56 & 79.63 \\
    TETFN${}^\dagger$   & 0.420 & 0.577 & 41.79 & 63.24 & \textbf{81.18} & 80.24 & 0.310 & 0.695 & 54.47 & 73.65 & 79.73 & 79.81 \\
    ALMT${}^*$          & \textbf{0.408} & 0.594 & 43.11 & 65.86 & 78.77 & 78.71 & 0.308 & 0.700 & 52.90 & 71.86 & 79.59 & 79.51 \\
    TMBL${}^*$          & 0.429 & 0.592 & 41.58 & 65.43 & 79.12 & 78.75 & 0.313 & 0.706 & 52.03 & 73.02 & 80.46 & 80.36 \\
    \hline
    \textbf{KuDA} & \textbf{0.408} & \textbf{0.613} & \textbf{43.54} & \textbf{66.52} & 80.74 & \textbf{80.71} & \textbf{0.271} & \textbf{0.759} & \textbf{61.22} & \textbf{76.21} & \textbf{82.11} & \textbf{82.04} \\
    \hline
  \end{tabular}
  \caption{Performance comparison on CH-SIMS and CH-SIMSv2. Note: the best result is marked in bold; ${}^\dagger$ means the result is from \cite{MaoYXYLG22}; ${}^*$ denotes the results are reproduced from code provided by their authors.}
  \label{tab:sims_experience}
\end{table*}

\begin{table*}[ht]\small
  \setlength{\tabcolsep}{5.2pt}
  \centering
  \begin{tabular}{l|ccccc|ccccc}
    \hline
    \multicolumn{1}{l|}{\multirow{2}{*}{\textbf{Methods}}} & \multicolumn{5}{c|}{\textbf{MOSI}} & \multicolumn{5}{c}{\textbf{MOSEI}} \\
    {} & \textbf{MAE} & \textbf{Corr} & \textbf{Acc-7} & \textbf{Acc-2} & \textbf{F1} & \textbf{MAE} & \textbf{Corr} & \textbf{Acc-7} & \textbf{Acc-2} & \textbf{F1} \\
    \hline
    TFN${}^\dagger$     & 0.947 & 0.673 & 34.46 & 77.99/79.08 & 77.95/79.11 & 0.572 & 0.714 & 51.60 & 78.50/81.89 & 78.96/81.74 \\
    LMF${}^\dagger$     & 0.950 & 0.651 & 33.82 & 77.90/79.18 & 77.80/79.15 & 0.576 & 0.717 & 51.59 & 80.54/83.48 & 80.94/83.36 \\
    MulT${}^\dagger$    & 0.879 & 0.702 & 36.91 & 79.71/80.98 & 79.63/80.95 & 0.559 & 0.733 & 52.84 & 81.15/84.63 & 81.56/84.52 \\
    MISA${}^\dagger$    & 0.776 & 0.778 & 41.37 & 81.84/83.54 & 81.82/83.58 & 0.557 & 0.751 & 52.05 & 80.67/84.67 & 81.12/84.66 \\
    BBFN${}^*$          & 0.796 & 0.744 & 43.88 & 80.32/82.47 & 80.21/82.44 & 0.545 & 0.760 & 52.88 & 82.87/85.73 & 83.13/85.56 \\
    MMIM${}^*$          & 0.744 & 0.780 & 44.75 & 82.51/84.30 & 82.38/84.23 & 0.550 & 0.761 & 51.88 & 83.75/85.42 & \textbf{83.93}/85.26 \\
    Self-MM${}^\dagger$ & 0.708 & \textbf{0.796} & 46.67 & 83.44/85.46 & 83.36/85.43 & 0.531 & 0.764 & \textbf{53.87} & \textbf{83.76}/85.15 & 83.82/84.90 \\
    CubeMLP${}^*$       & 0.755 & 0.772 & 43.44 & 80.76/82.32 & 81.77/84.23 & 0.537 & 0.761 & 53.35 & 82.36/85.23 & 82.61/85.04 \\
    ALMT${}^*$          & 0.712 & 0.793 & 46.79 & 83.97/85.82 & 84.05/85.86 & 0.530 & 0.774 & 53.62 & 81.54/85.99 & 81.05/86.05 \\
    \hline
    \textbf{KuDA}  & \textbf{0.705} & 0.795 & \textbf{47.08} & \textbf{84.40}/\textbf{86.43} & \textbf{84.48}/\textbf{86.46} & \textbf{0.529} & \textbf{0.776} & 52.89 & 83.26/\textbf{86.46} & 82.97/\textbf{86.59} \\
    \hline
  \end{tabular}
  \caption{Performance comparison on MOSI and MOSEI. Note: the best result is highlighted in bold; in Acc-2 and F1, the left of the / corresponds to ``negative/non-negative'' and the right corresponds to ``negative/positive''; ${}^\dagger$ means the result is from \citep{MaoYXYLG22}; ${}^*$ denotes the results are reproduced from code provided by their authors.}
  \label{tab:mos_experience}
\end{table*}

\subsection{Performance Comparison}
\label{sec:PerformanceComparison}
Table~\ref{tab:sims_experience} and Table~\ref{tab:mos_experience} present the comparison results of the baselines and the proposed method on CH-SIMS, CH-SIMSv2, MOSI and MOSEI.

\begin{table}[ht]\footnotesize
  \setlength{\tabcolsep}{3pt}
  \centering
  \begin{tabular}{l|cccccc}
    \hline
    \multicolumn{1}{l|}{\multirow{2}{*}{Methods}} & \multicolumn{3}{c}{CH-SIMSv2} & \multicolumn{3}{c}{MOSI} \\
    {} & MAE & Corr & Acc-5 & MAE & Corr & Acc-7 \\
    \hline
    \textbf{KuDA} & \textbf{0.271} & \textbf{0.759} & \textbf{61.22} & \textbf{0.705} & 0.795 & \textbf{47.08} \\
    \hline
    w/o KIP     & 0.288 & 0.729 & 56.87 & 0.731 & 0.798 & 44.61 \\
    w/o Adapter & 0.286 & 0.735 & 57.54 & 0.714 & 0.787 & 46.79 \\
    w/o EKI      & 0.293 & 0.733 & 56.48 & 0.742 & 0.778 & 44.46 \\
    \hline
    w/o SR     & 0.281 & 0.736 & 58.12 & 0.729 & 0.798 & 45.19 \\
    w/o DAF     & 0.309 & 0.716 & 54.06 & 0.754 & 0.779 & 43.73 \\
    w/o CE Loss & 0.277 & 0.752 & 59.38 & 0.712 & \textbf{0.799} & 44.31 \\
    \hline
  \end{tabular}
  \caption{Ablation results of KuDA's components on CH-SIMSv2 and MOSI. Note: ``KIP'' is the Knowledge Inject Pretraining; ``EKI'' is denoted the Encoding with Knowledge Injection module; ``SR'' denotes the Sentiment Ratio; ``DAF'' denotes the Dynamic Attention Fusion module; the best result is highlighted in bold.}
  \label{tab:ablation_study}
\end{table}

Since the distribution of modality importance in the CH-SIMS and CH-SIMSv2 is more uniform, they are more complex than MOSI and MOSEI. As shown in Table~\ref{tab:sims_experience}, the proposed method outperforms all baselines on all metrics. It is worth noting that our method achieves superior performance on the CH-SIMSv2 dataset. For example, compared with ALMT (text center) and TMBL (ternary symmetric), our method achieves 8.32\% and 9.19\% improvement on the Acc-5 and also achieves significant improvement on the Acc-3. Therefore, achieving superior performance in the more challenging scenario indicates that KuDA can adjust the distribution of modality weights to complete dynamic fusion. It also shows that adjusting the dominant modality is crucial for MSA. Furthermore, although the importance of modalities is not evenly distributed in MOSI and MOSEI, and the text modality plays an important role, it can be seen from Table~\ref{tab:mos_experience} that KuDA still obtained SOTA performance in almost all metrics. At the same time, KuDA surpasses some text-center training methods, such as BBFN and ALMT.

\subsection{Ablation Study and Analysis}
\label{sec:AblationStudyandAnalysis}

\subsubsection{Effects of Different Components}
\label{EffectsofDifferentComponents}
We conducted ablation studies to validate the effectiveness of each component, as shown in Table~\ref{tab:ablation_study}. By comparing the ``w/o DAF'' and KuDA, we observe that removing the DAF can seriously reduce performance. This means that KuDA dynamically adjusts the attention weights between modalities for different scenarios to select the dominant modality. In addition, the performance decreases in ``w/o EKI'', which shows that sentiment knowledge can further guide dynamic fusion. The performance decreases after removing other modules, showing their effectiveness. Since improving the utilization of vision and audio on MOSI introduces noise, Corr has improved.

\begin{table}[ht]\footnotesize
  \renewcommand{\arraystretch}{1.1}
  \setlength{\tabcolsep}{3.8pt}
  \centering
  \begin{tabular}{c|cccccc}
    \hline
    \multicolumn{1}{c|}{\multirow{2}{*}{Methods}} & \multicolumn{3}{c}{CH-SIMSv2} & \multicolumn{3}{c}{MOSI} \\
    {} & MAE & Corr & Acc-5 & MAE & Corr & Acc-7 \\
    \hline
    V+A  & 0.362 & 0.562 & 47.20 & 1.370 & 0.235 & 17.20 \\
    T+V  & 0.298 & 0.721 & 54.93 & 0.769 & 0.770 & 43.29 \\
    T+A  & 0.335 & 0.645 & 52.13 & 0.733 & \textbf{0.795} & 42.27 \\
    \hline
    \rowcolor{gray!30}
    \textbf{KuDA} & \textbf{0.271} & \textbf{0.759} & \textbf{61.22} & \textbf{0.705} & \textbf{0.795} & \textbf{47.08} \\
    \hline
    \hline
    V+A  & 0.451 & 0.449 & 39.94 & 1.437 & 0.201 & 15.74 \\
    T+V  & 0.346 & 0.623 & 48.26 & 0.772 & 0.778 & 43.15 \\
    T+A  & 0.368 & 0.586 & 46.03 & 0.736 & 0.788 & 43.88 \\
    \hline
    \rowcolor{gray!30}
    \textbf{ALMT} & \textbf{0.308} & \textbf{0.700} & \textbf{52.90} & \textbf{0.712} & \textbf{0.793} & \textbf{46.79} \\
    \hline
    \hline
    V+A  & 0.452 & 0.361 & 38.97 & 1.453 & 0.137 & 15.45 \\
    T+V  & 0.354 & 0.615 & 48.94 & 0.796 & 0.745 & 41.98 \\
    T+A  & 0.362 & 0.623 & 47.87 & 0.816 & 0.739 & 41.54 \\
    \hline
    \rowcolor{gray!30}
    \textbf{CubeMLP} & \textbf{0.334} & \textbf{0.648} & \textbf{52.90} & \textbf{0.755} & \textbf{0.772} & \textbf{43.44} \\
    \hline
    \hline
  \end{tabular}
  \caption{Importance of different modalities on KuDA, ALMT (text center) and CubeMLP (ternary symmetric). T, V, and A represent text, vision, and audio modalities. Note: the best result is highlighted in bold.}
  \label{tab:important_modality}
\end{table}

\begin{table}[htp]\scriptsize
  \renewcommand{\arraystretch}{1.3}
  \setlength{\tabcolsep}{3.6pt}
  \centering
  \begin{tabular}{c|cccccc}
    \hline
    Fusion Methods & MAE & Corr & Acc-5 & Acc-3 & Acc-2 & F1 \\
    \hline
    Concatenation                        & 0.296 & 0.728 & 79.98 & 72.53 & 54.64 & 79.93 \\
    Addition                             & 0.304 & 0.721 & 78.14 & 72.24 & 54.55 & 78.02 \\
    Tensor Fusion (TFN)                  & 0.282 & 0.754 & 80.75 & 75.05 & 55.80 & 80.64 \\
    Low-rank Fusion (LMF)                & 0.321 & 0.711 & 79.50 & 72.24 & 53.09 & 79.48 \\
    CMT (BBFN, ALMT) & 0.282 & 0.745 & 81.43 & 74.08 & 57.16 & 81.33 \\
    \hline
    \textbf{KuDA} & \textbf{0.271} & \textbf{0.759} & \textbf{82.11} & \textbf{76.21} & \textbf{61.22} & \textbf{82.04} \\
    \hline
  \end{tabular}
  \caption{The performance of different fusion methods on CH-SIMSv2. Note: the CMT denotes Cross-modal Transformer.}
  \label{tab:fusion_method}
\end{table}

\subsubsection{Importance of Different Modalities}
\label{ImportanceofDifferentModalities}
To validate the impact of text, vision and audio modalities, we performed ablation studies that removed each modality on KuDA, ALMT and CubeMLP, as shown in Table~\ref{tab:important_modality}.

In CH-SIMSv2, which has more complex scenes, we can see that KuDA can reach the SOTA when each modality is removed. Meanwhile, when performance degrades, KuDA can still achieve acceptable results. In contrast, the other baselines will have a significant performance degradation, which proves that our method can adaptive focus on the suboptimal modality to capture sentiment features. For the MOSI dataset, which is mainly text, we can see that CubeMLP has dropped significantly when each modality is removed. Furthermore, ALMT drops sharply after removing text and is lower than KuDA. Notably, observing the performance degradation trend after removing a certain modality indicates that the importance of each modality is evenly distributed in the CH-SIMSv2, while the importance of text modality is higher in MOSI.

\subsubsection{Effects of Different Fusion Methods}
\label{EffectsofDifferentFusionMethods}
To analyze the effects of different fusion techniques, we conducted some experiments shown in Table~\ref{tab:fusion_method}. Obviously, when faced with complex scenes, using either ternary symmetric-based (TFN, LMF) or text center-based (BBFN, ALMT) fusion methods will result in performance decline. This indicates that not focusing on the dominant modality or statically setting the dominant modality will limit the performance of MSA. However, the use of our Dynamic Attention Fusion to dynamically fuse unimodal features is the most effective.

\begin{figure}[ht]
  \centering
  \subfloat[\centering change on CH-SIMSv2 \label{fig:alpha_chsimsv2}]{\includegraphics[width=0.5\linewidth]{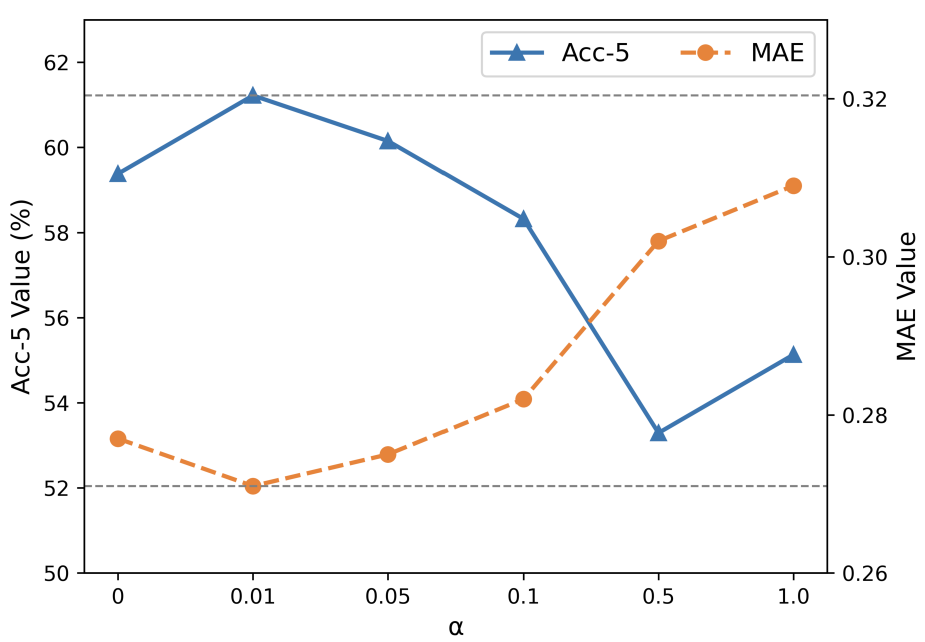}}
  \hfil
  \subfloat[\centering change on MOSI \label{fig:alpha_mosi}]{\includegraphics[width=0.5\linewidth]{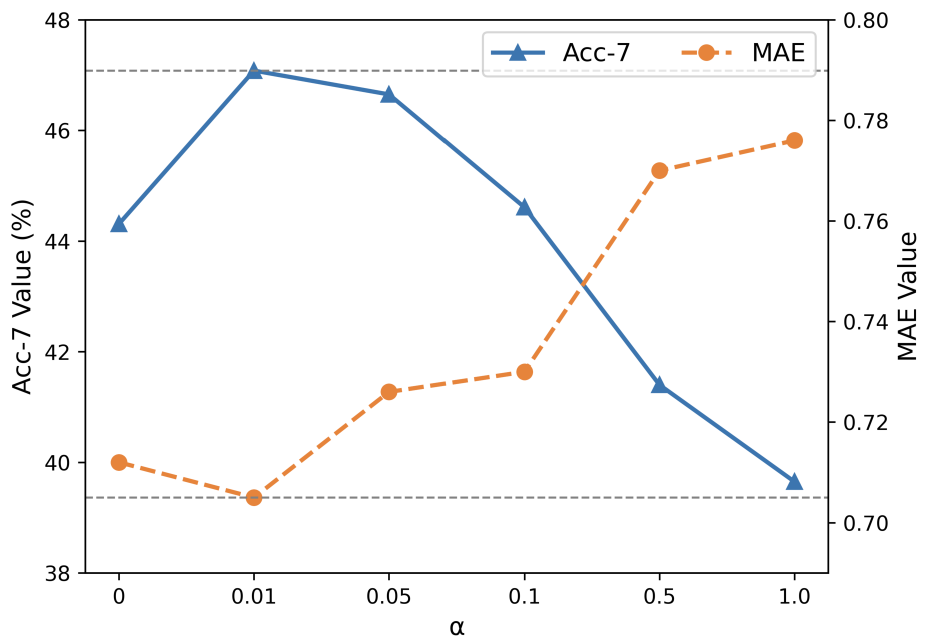}}
  \caption{Visualization of performance with change $\alpha$ on CH-SIMSv2 and MOSI.}
  \label{fig:alpha}
\end{figure}

\begin{figure}[htp]
  \centering
  \subfloat[CubeMLP \label{fig:tsne_cubemlp}]{\includegraphics[width=0.5\linewidth]{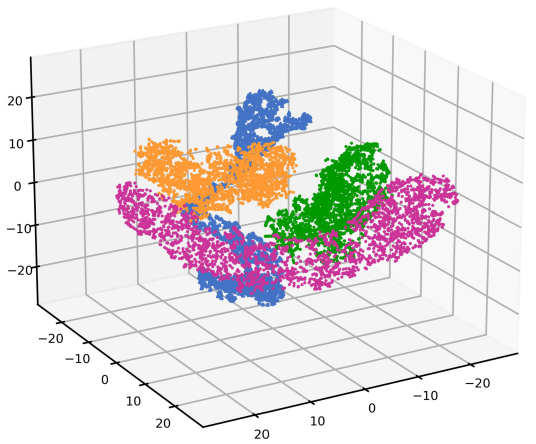}}
  \hfil
  \subfloat[ALMT \label{fig:tsne_almt}]{\includegraphics[width=0.5\linewidth]{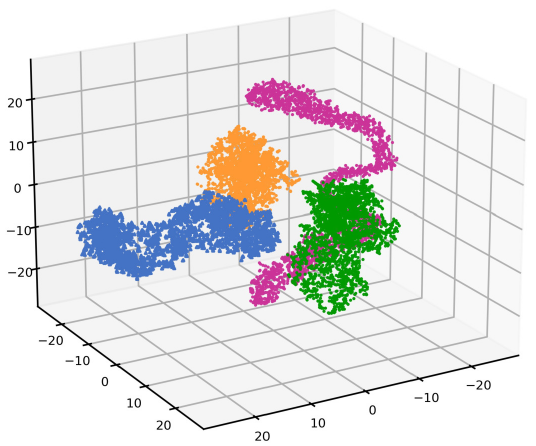}}

  \subfloat[KuDA \label{fig:tsne_kuda}]{\includegraphics[width=0.5\linewidth]{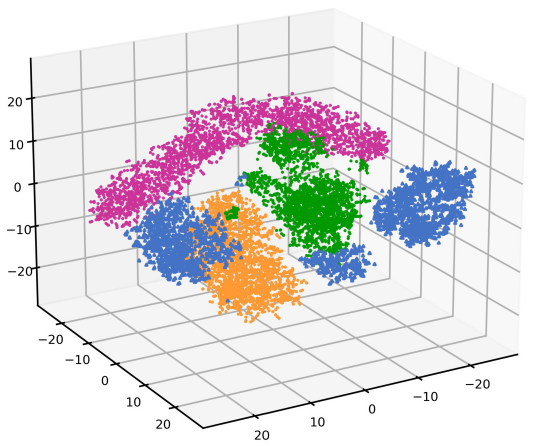}}
  \hfil
  \subfloat{\includegraphics[width=0.5\linewidth]{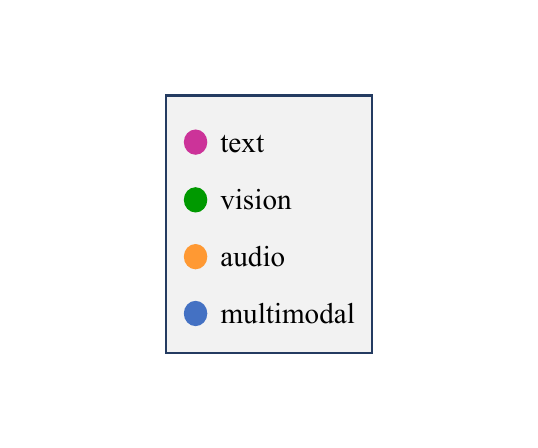}\label{fig:legend}}
  \caption{The t-SNE visualization \citep{vandermaaten08a} of the unimodal features (text, vision, and audio) and multimodal features in (a) CubeMLP; (b) ALMT; (c) KuDA.}
  \label{fig:tsne}
\end{figure}

\subsubsection{Effects of Correlation Estimation}
\label{EffectsofMulti-TaskLearning}
As shown in Figure~\ref{fig:alpha}, we discuss the impact of CE loss on CH-SIMSv2 and MOSI by modifying the $\alpha$. We compare the MAE, Acc-5 and Acc-7 as these metrics indicate the method's ability to predict fine-grained sentiment.

Compared to removing CE loss, i.e. $\alpha$=0, the model's performance achieves STOA when using CE loss and setting $\alpha$=0.01. This shows that CE loss can highlight the contribution of the dominant modality further. However, the performance shows a downward trend when $\alpha$ increases, indicating that KuDA will enhance the retention of non-dominant modality features in the multimodal representation when CE loss increases, limiting its performance.

\begin{table*}[ht]\scriptsize
  \renewcommand{\arraystretch}{1.2}
  \setlength{\tabcolsep}{2.5pt}
  \centering
  \begin{tabular}{p{1cm}|m{4.75cm}<{\centering}|m{4.75cm}<{\centering}|m{4.75cm}<{\centering}}
    \hline
    \textbf{Cases}   & \textbf{Case (a)}  & \textbf{Case (b)}  & \textbf{Case (c)} \\
    \hline
    \textcolor[RGB]{0,153,0}{\textbf{Vision}}  & {\includegraphics[scale=0.15]{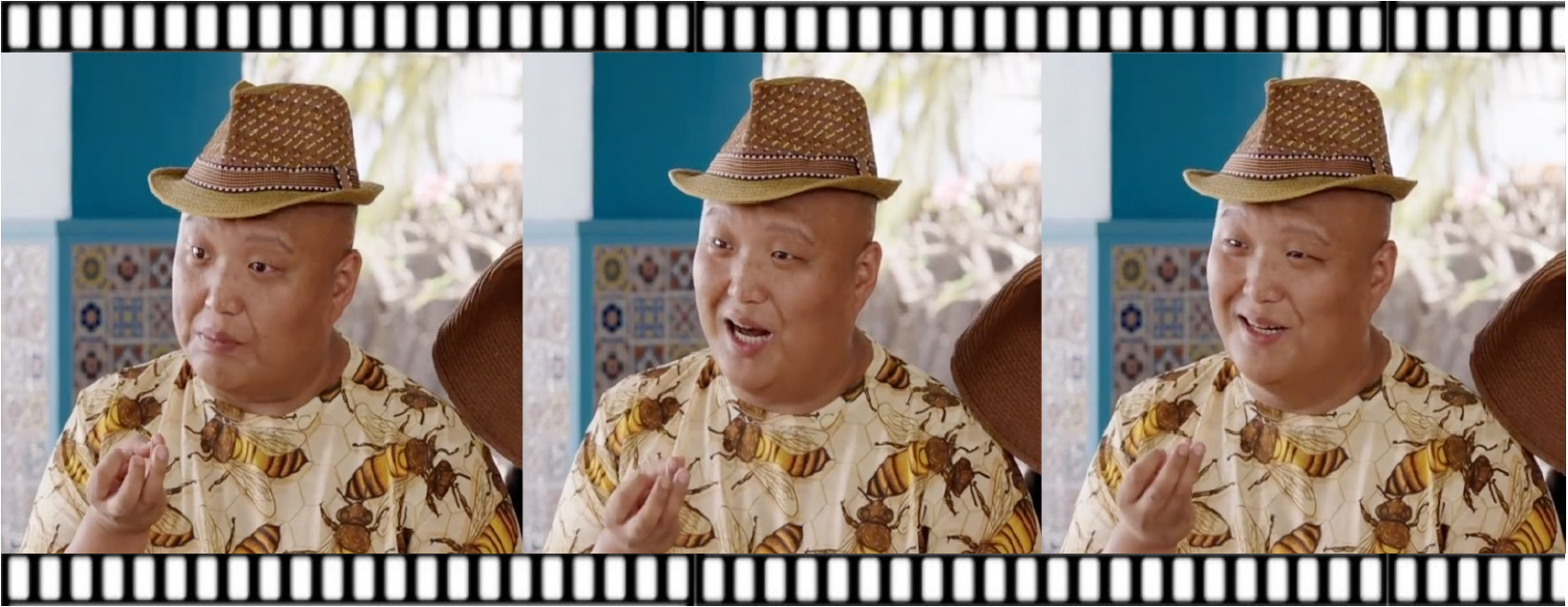}} & {\includegraphics[scale=0.15]{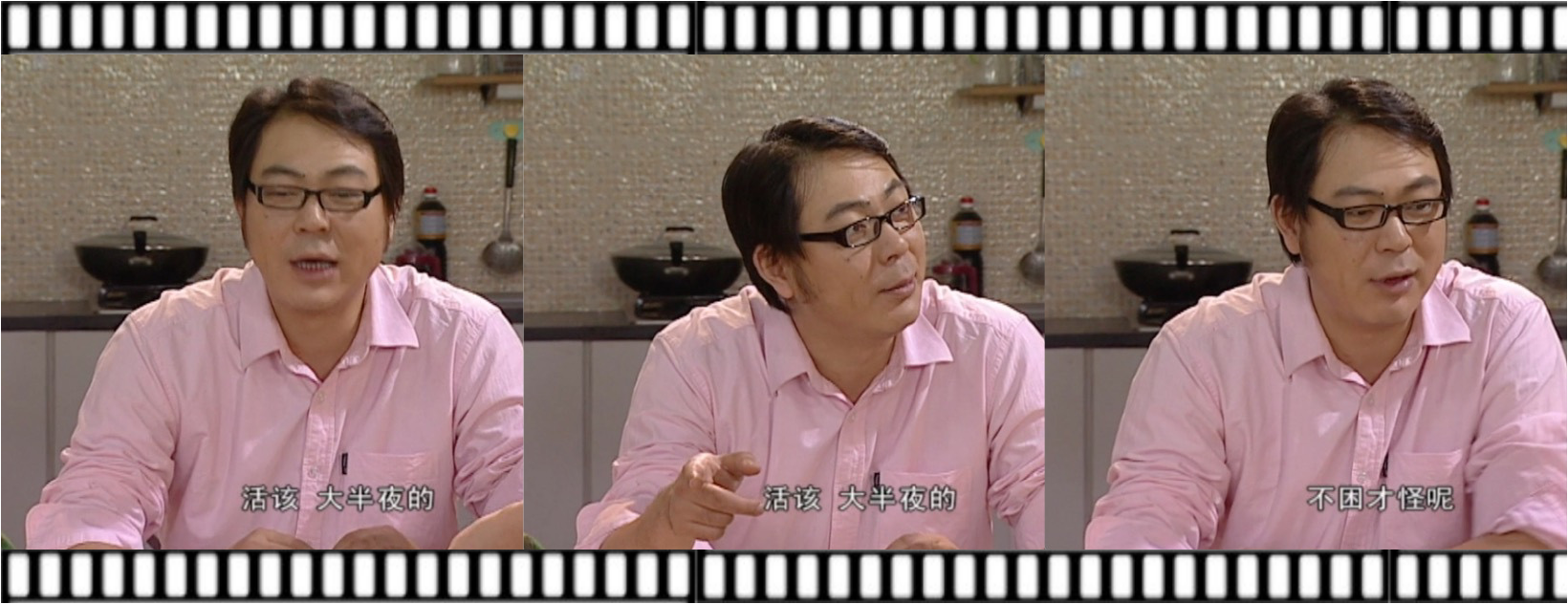}} & {\includegraphics[scale=0.15]{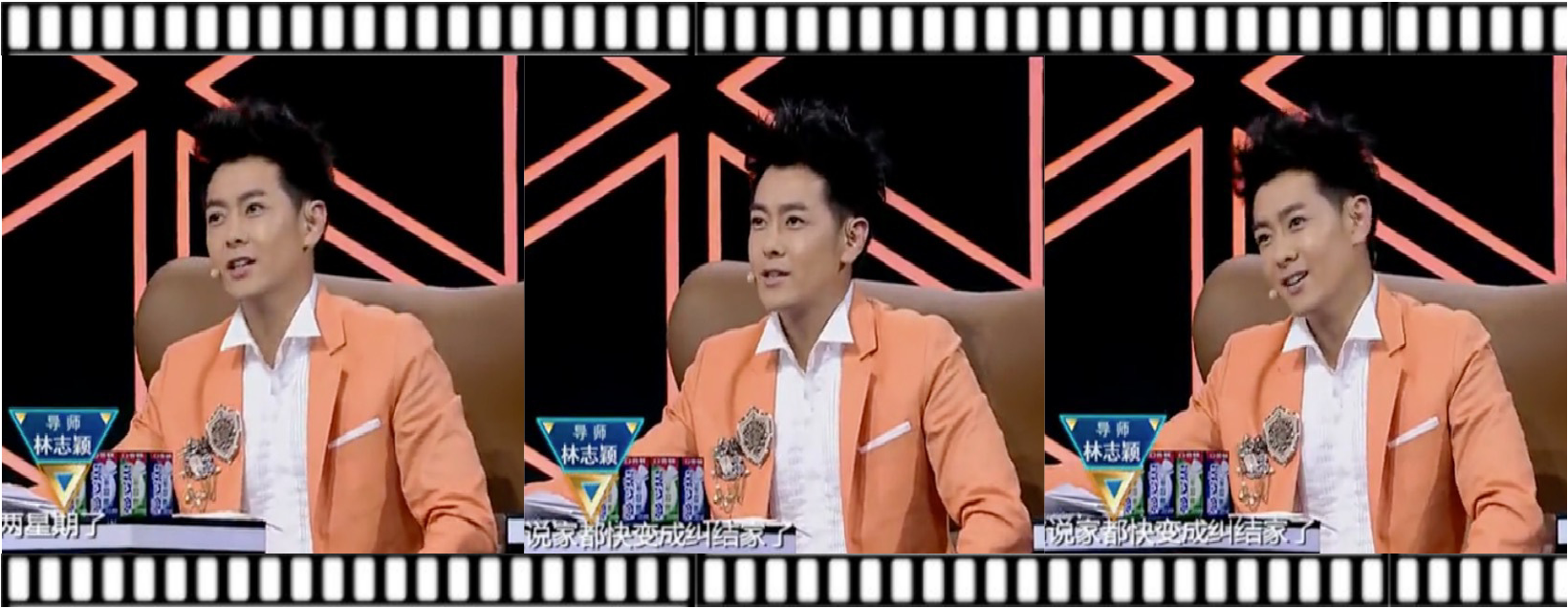}} \\
    \textcolor[RGB]{204,51,153}{\textbf{Text}}    & \textbf{Aren't you almost bankrupt?} & \textbf{He deserves it, he must be very sleepy!} & \textbf{You have been struggling for two weeks.} \\
    \textcolor[RGB]{255,153,51}{\textbf{Audio}} & {\includegraphics[scale=0.15]{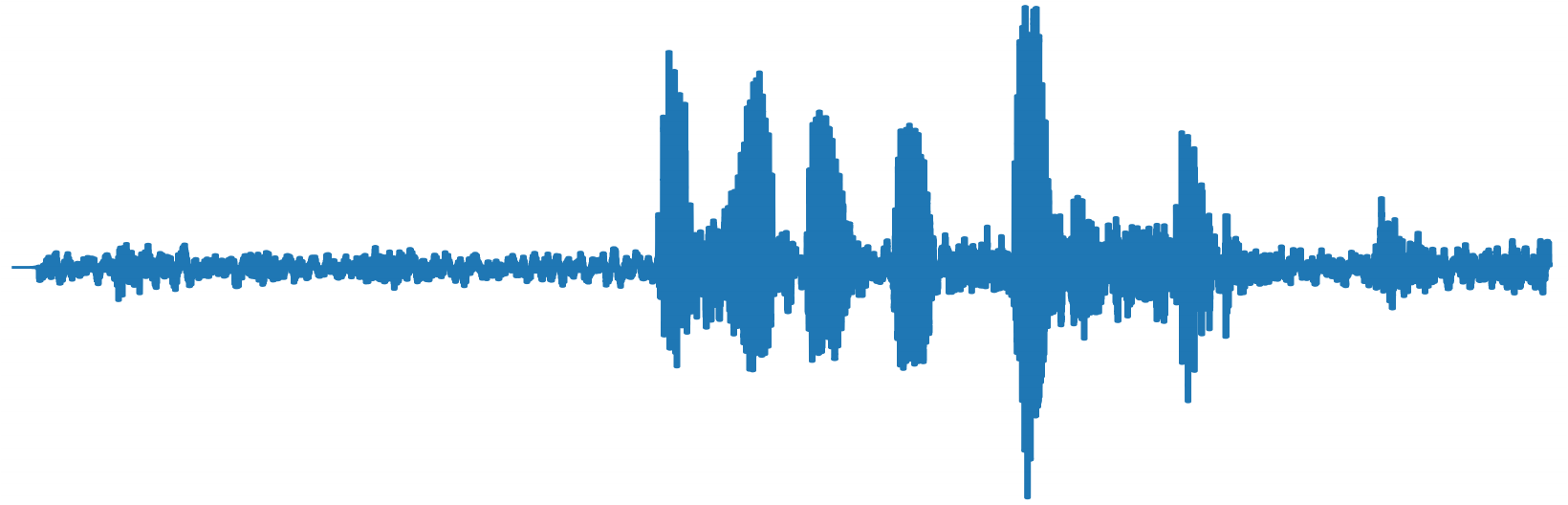}} & {\includegraphics[scale=0.15]{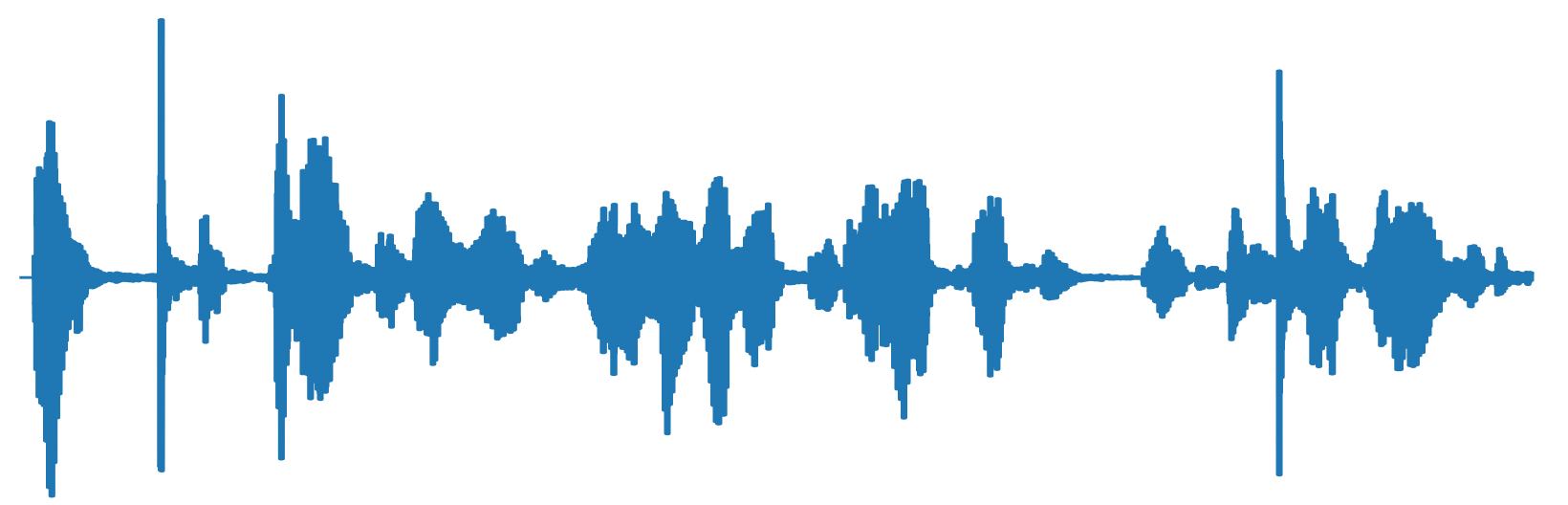}} & {\includegraphics[scale=0.15]{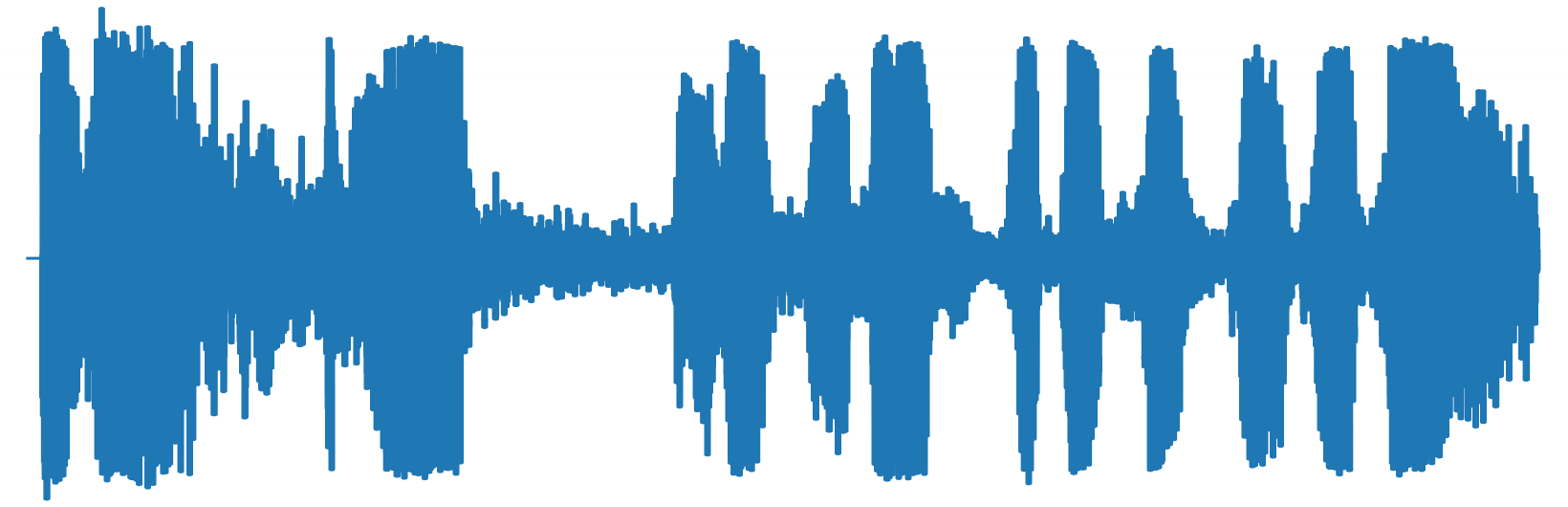}} \\
    \hline
    \tabincell{l}{\textbf{Ground}\\\textbf{Truth}} & \tabincell{c}{\textbf{V: 1.0 \qquad \ T: -0.8 \qquad \ A: -0.8}\\\textbf{M = 0.6}} & \tabincell{c}{\textbf{V: 0.6 \qquad \ T: -1.0 \qquad \ A: 0.6}\\\textbf{M = -0.8}} & \tabincell{c}{\textbf{V: 0.8 \qquad \ T: -0.8 \qquad \ A: 0.0}\\\textbf{M = 0.0}} \\
    \hline
    \textbf{Output} & \textbf{0.4} & \textbf{-1.0} & \textbf{0.0} \\
    \tabincell{l}{\textbf{Attention}\\\textbf{Weight}} & {\includegraphics[scale=0.135]{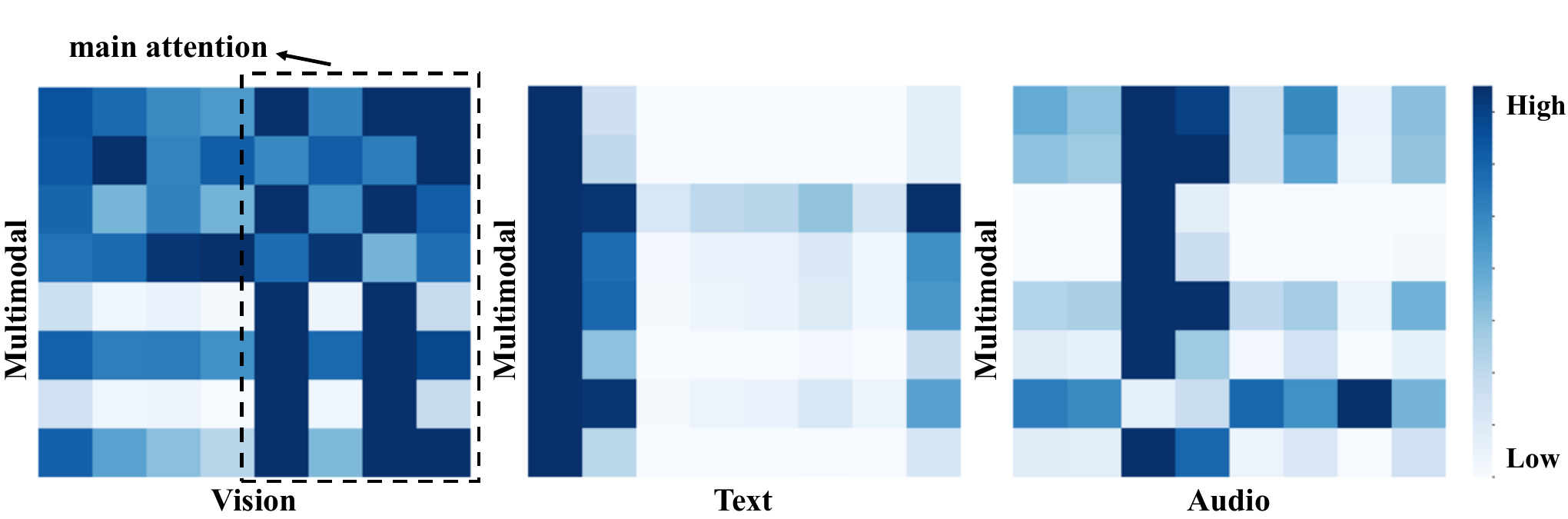}} & {\includegraphics[scale=0.135]{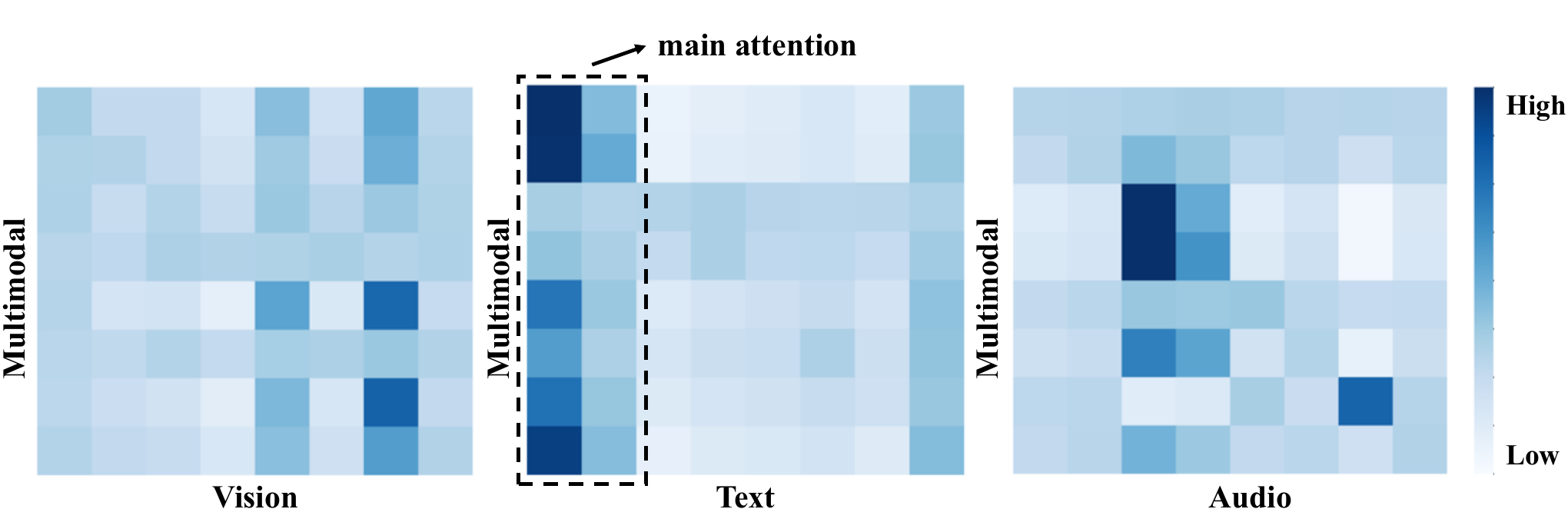}} & {\includegraphics[scale=0.135]{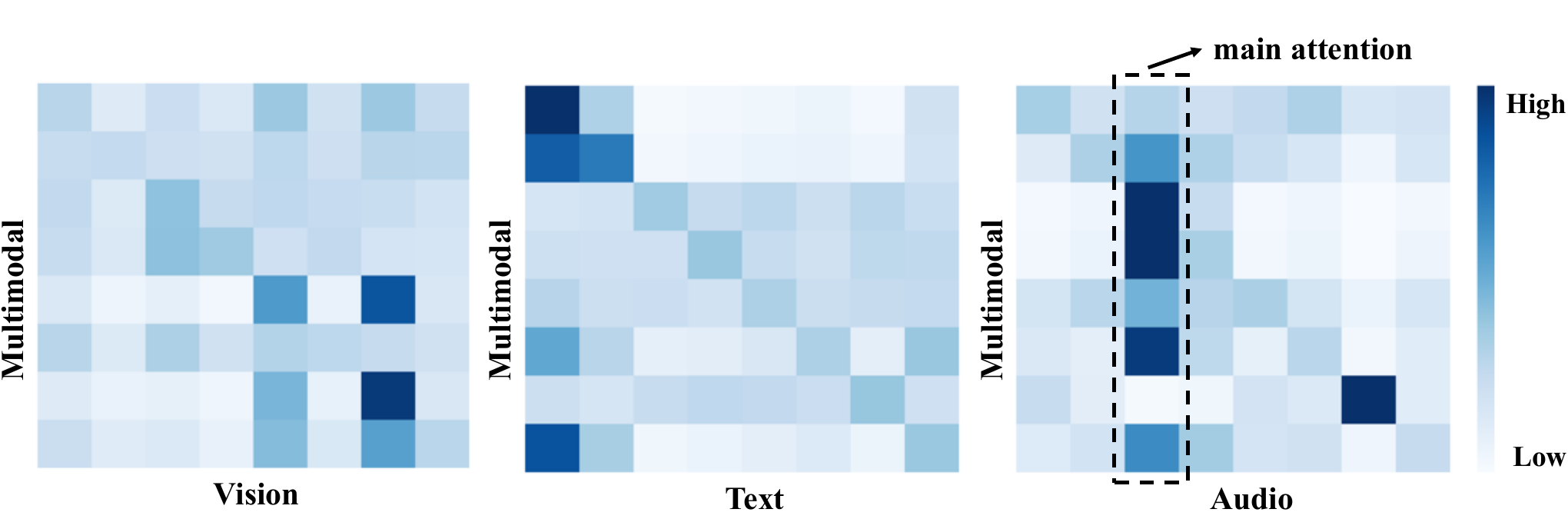}} \\
    \hline
  \end{tabular}
  \caption{Three cases from CH-SIMS. Ground Truth consists of unimodal labels (V, T, and A) and the multimodal label (M). Output shows the multimodal prediction from KuDA. Attention Weight shows the distribution of attention weights from the multimodal features with the features of vision, text, and audio modalities. In Attention Weight, the left, middle, and right figures of each case represent the multimodal and vision, text, and audio modality weights, respectively. The partial attention of the dominant modality is marked with a dashed box.}
  \label{tab:case_study}
\end{table*}

\subsubsection{Visualization of Features Distribution}
\label{VisualizationofFeaturesDistribution}
To verify KuDA can dynamically select dominant modality, we use t-SNE to visualize the features of text, vision, audio and multimodal on CH-SIMSv2, as shown in Figure~\ref{fig:tsne}. We then selected two typical methods, CubeMLP (ternary symmetric) and ALMT (text center), to compare with KuDA.

In Figure~\ref{fig:tsne_cubemlp}, we can see that since CubeMLP treats contributions of each modality equally, all unimodal features are averaged around the multimodal features. In addition, as shown in Figure~\ref{fig:tsne_almt}, due to ALMT is a text center-based method, we can observe that the text features is in the middle of the audio and vision, and all unimodal features are distributed on the other side of multimodal features. However, as can be seen from Figure~\ref{fig:tsne_kuda}, the difference is that KuDA's multimodal features is divided into three clusters and is close to the text, audio, and vision features respectively. This indicates that KuDA dynamically selects the dominant modality to make the multimodal features closer to it.

\subsubsection{Case Study}
\label{CaseStudy}
To better prove that the our method can dynamically adjust contributions of different modalities, we selected three challenging cases for further analysis, as shown in Table~\ref{tab:case_study}.

We can observe that in case (a), although the text and audio express stronger negative sentiments, KuDA can still output the correct prediction. This case shows that by adjusting vision as the dominant modality, KuDA effectively captures the speaker's information of expression and action, which also guides the fusion of text and audio modalities. In cases (b) and (c), the similar distributions of labels also occurred. KuDA still makes correct predictions, which indicates that it captures the semantic information of text in case (b) and the intonation information of audio in case (c) by adjusting the attention weights. Meanwhile, it can be seen in the Attention Weight of Table~\ref{tab:case_study} that attention weight for the dominant modality (there are denser dark blocks) is higher than that for the other modalities. This once again proves the importance of dynamic attention fusion for the MSA task.

\section{Conclusion}
\label{sec:Conclusion}
In this paper, we propose a Knowledge-Guided Dynamic Modality Attention Fusion Framework (KuDA) to simultaneously solve the MSA task of the modality importance being equally or unequally distributed. Since KuDA dynamically adjusts the contribution of each modality for different scenarios, it effectively improves the utilization of the dominant modality. This enables our model to be more effective and generalized on four popular MSA benchmark datasets. At last, we perform comprehensive ablation studies to analyze this phenomenon.

\section*{Limitations}
Although the KuDA proposed in this paper has yielded exceptional outcomes, there remain several limitations that offer opportunities for further enhancement. First, KuDA suffers from an error propagation problem due to its two-stage training method. When the pretrained sentiment knowledge incorrectly predicts the sentiment score, the sentiment ratio will introduce noise when the model adjusts the weights. Second, KuDA needs to be pretrained using the sentiment knowledge of each modality, which increases the resource consumption of model training. In future work, we will further try to explore fine-tuning the pretrained knowledge injection module in the prediction stage to solve the above limitations.

\section*{Acknowledgements}
This work was supported by National Natural Science Foundation of China (Nos. 62062027, 62362015 and U22A2099), Innovation Project of GUET Graduate Education (No. 2024YCXS042).

\bibliography{custom}

\appendix
\section{Data Statistics and Analysis}
\label{sec:DataInvestigationandStatistics}
To explore the sample distribution that each modality as dominant and verify our idea, we investigated four MSA benchmark datasets (MOSI, MOSEI, CH-SIMS, and CH-SIMSv2). The statistical and analytical results are as follows.

\begin{figure}[ht]
  \includegraphics[width=\columnwidth]{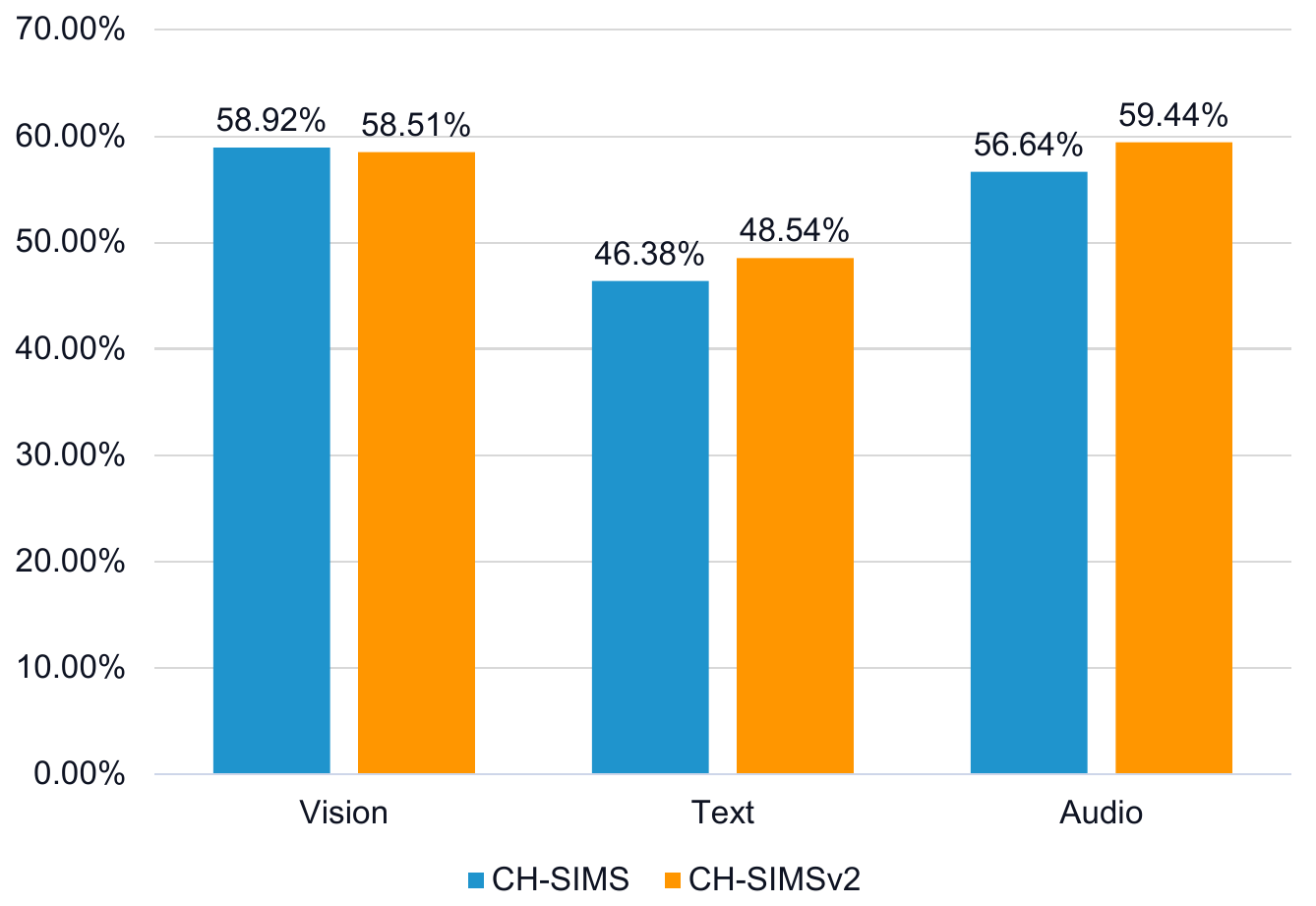}
  \caption{The distribution where any modality as dominant.}
  \label{fig:dominant}
\end{figure}

\begin{figure}[ht]
  \includegraphics[width=\columnwidth]{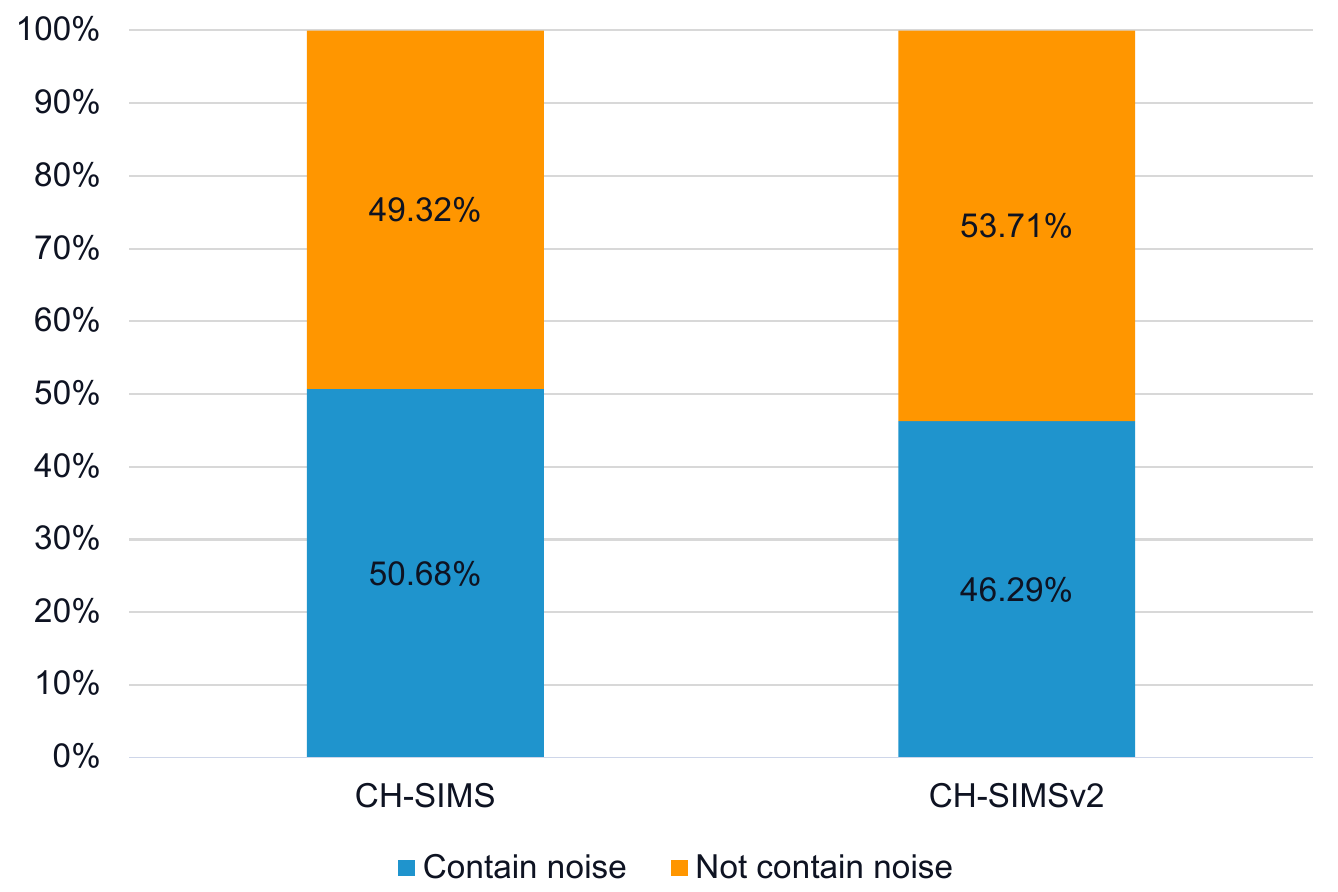}
  \caption{The distribution where the sample contains the noise modality.}
  \label{fig:noise}
\end{figure}

For MOSI and MOSEI, some researchers \citep{yuetal2020ch,liu2022make} have shown that the text modality is of higher importance and dominant. At the same time, some ablation studies \citep{hazarika2020misa,lin2022multimodal,yu2023conki} report about a 30\% binary accuracy drop in these two datasets when removing the text modality (80\%+ with text, while about 50\%+ without text). Thus, the majority of samples in MOSI and MOSEI use text as the dominant modality.

\begin{algorithm}[ht]\small
  \SetAlgoLined
  \caption{Training Process of KuDA}\label{agr:process}
    \KwSty{Stage 1: Knowledge Inject Pretraining} \\
    \KwIn{External dataset $\varphi$ with the unimodal features $I_{m}$ and labels $y_{m},m \in \left\{t, v, a\right\}$}
    \KwOut{Pretrained adapters and decoders $\{ \theta_{m}^{adapter}, \theta_{m}^{decoder} | m \in \left \{t, v, a \right \} \}$}
      \For{each training epoch}{
        \For{batch $\left \{ \left ( I_{t}^{i},I_{v}^{i},I_{a}^{i}  \right )  \right \}_{i=1}^{N}$ from $\varphi$}{
          Encode $I_{m}^{i}$ to $U^{i}_{m}$ as Eq.~(\ref{eq:bertencoder})-(\ref{eq:knowenhance}) \\
          Predict $\hat{y}^i_m$ using decoders as Eq.~(\ref{eq:decoder}) \\
          Compute $\mathcal{L}_{reg}$ of each modality with $\hat{y}^i_m$ and $y^{i}_{m}$ as Eq.~(\ref{eq:loss_reg}) \\
          Update parameters of encoding with knowledge injection module $\{ \theta_{m}^{know} | m \in \left \{t, v, a \right \} \}$
        }
        Save $\{ \theta_{m}^{adapter}, \theta_{m}^{decoder} | m \in \left \{t, v, a \right \} \}$ when achieves the best validate results
      }
    \BlankLine
    \KwSty{Stage 2: Downstream Training} \\
    \KwIn{Target dataset $\mathcal{D}$ with the features $I_{m}, m \in \left\{t, v, a\right\}$ and labels $y$; Pretrained adapters and decoders}
    \KwOut{Predictions $\hat{y}$}
      \For{each training epoch}{
        \For{batch $\left \{ \left ( I_{t}^{i},I_{v}^{i},I_{a}^{i}  \right )  \right \}_{i=1}^{N}$ from $\mathcal{D}$}{
          Encode $I_{m}^{i}$ to $U^{i}_{m}$ as Eq.~(\ref{eq:bertencoder})-(\ref{eq:knowenhance}) \\
          Predict $\hat{y}^i_m$ using decoders and calculate $R^{i}_{m}$ as Eq.~(\ref{eq:decoder}),~(\ref{eq:unisentiratio}) \\
          Process dynamic attention fusion as Eq.~(\ref{eq:firstblock})-(\ref{eq:fusion2}) \\
          Predict $\hat{y}^i$ using MLP as Eq.~(\ref{eq:prediction}) \\
          Compute $\mathcal{L}_{cor}$, $\mathcal{L}_{reg}$, $\mathcal{L}_{task}$ as Eq.~(\ref{eq:cor_loss}),~(\ref{eq:loss_reg}),~(\ref{eq:loss_task}) \\
          Update the model parameters except $\{ \theta_{m}^{adapter}, \theta_{m}^{decoder} | m \in \left \{t, v, a \right \} \}$
        }
      }
\end{algorithm}

For CH-SIMS and CH-SIMSv2, we used the unimodal labels provided by the dataset to statistic the number of samples, as shown in Figure~\ref{fig:dominant}. We can observe that the number of samples dominated by vision, text, and audio modalities accounts for 45\%-60\% of the total, and the distribution of each modality is even. This indicates that each modality will be dominant, and this situation is not uncommon. It also shows that it is necessary to adjust the dominant modality dynamically. Notably, the sum of the proportions of three modalities in the same dataset is more than 100\% because there may be multiple modalities that dominate in the same sample. Then, we statistic the number of samples containing noise modality (the sentiment polarity of this unimodal is different from that of the multimodal), as shown in Figure~\ref{fig:noise}. We can see that the proportion of samples containing noise modality is around 50\%, which further shows that the inability to dynamically adjust the contribution of each modality will limits the performance of MSA.

\section{Training Process}
\label{sec:TrainingProcess}
KuDA uses a two-stage training method, which details are shown in Algorithm~\ref{agr:process}. In Stage 1, we pretrained the Encoding with Knowledge Injection module using external data. For CH-SIMS and CH-SIMSv2, we use the unimodal labels of the dataset itself for pretraining. For MOSI and MOSEI, considering the data scale, we translate the texts of CH-SIMS and CH-SIMSv2 into English and inject the sentiment knowledge of CH-SIMS into MOSI and that of CH-SIMSv2 into MOSEI. Notably, to compare fairly with the baselines, we only pretrain the task of unimodal sentiment prediction on all datasets and do not involve the MSA task. In Stage 2, to prevent the pretrained knowledge from being overwritten, we froze the Adapter and Decoder of each modality and performed the MSA task based on the pretrained knowledge.

\end{document}